\documentclass[lettersize,journal]{IEEEtran}
\usepackage{amsmath,amsfonts}
\usepackage{array}
\usepackage{cite}
\usepackage{textcomp}
\usepackage{stfloats}
\usepackage{url}
\usepackage{algorithm2e}
\usepackage{algpseudocode}
\usepackage{verbatim}
\usepackage{float}
\usepackage{multirow}
\usepackage{diagbox}
\usepackage{ulem}
\usepackage{graphicx}
\usepackage{xcolor}
\usepackage{cite}
\usepackage{bbding}
\usepackage{wrapfig}
\usepackage{fontawesome}
\usepackage{booktabs}
\usepackage[colorlinks,linkcolor=red,citecolor=green]{hyperref}
\usepackage{soul}
\usepackage{bm}

\usepackage[caption=false,font=footnotesize,labelfont=rm,textfont=rm]{subfig}

\newcommand{\revise}[1]{\textcolor{black}{#1}}
\newcommand{\revisesec}[1]{\textcolor{black}{#1}}

\begin{document}

\title{Human as Points: Explicit Point-based 3D Human Reconstruction from Single-view RGB Images}

\author{Yingzhi~Tang,~Qijian~Zhang, Yebin Liu, and ~Junhui Hou 
\thanks{Y. Tang, Q. Zhang, and J. Hou are with the Department of Computer Science, City University of Hong Kong, Hong Kong SAR. Email: yztang4-c@my.cityu.edu.hk; qijizhang3-c@my.cityu.edu.hk; jh.hou@cityu.edu.hk.} 
\thanks{Y. Liu is with the Department of Automation,
Tsinghua University, Beijing, China. Email: liuyebin@mail.tsinghua.edu.cn}
\thanks{This work was supported in part by the NSFC Excellent Young Scientists Fund 62422118, and in part by the Hong Kong Research Grants Council under Grants 11219324, 11202320, and 11219422. \textit{Corresponding author: Junhui Hou}}
}

\markboth{Revised Manuscript Submitted to IEEE TPAMI}%
{Shell \MakeLowercase{\textit{et al.}}: A Sample Article Using IEEEtran.cls for IEEE Journals}


\maketitle

\begin{abstract}
    The latest trends in the research field of single-view human reconstruction are devoted to learning deep implicit functions constrained by explicit body shape priors. Despite the remarkable performance improvements compared with traditional processing pipelines, existing learning approaches still exhibit limitations in terms of \textit{flexibility}, \textit{generalizability}, \textit{robustness}, and/or \textit{representation capability}. To comprehensively address the above issues, in this paper, we investigate an explicit point-based human reconstruction framework named HaP, which utilizes point clouds as the intermediate representation of the target geometric structure. Technically, our approach features fully explicit point cloud estimation \revise{(exploiting depth and SMPL)}, manipulation \revise{(SMPL rectification)}, generation \revise{(built upon diffusion)}, and refinement \revise{(displacement learning and depth replacement)} in the 3D geometric space, instead of an implicit learning process that can be ambiguous and less controllable. Extensive experiments demonstrate that our framework achieves quantitative performance improvements of 20$\%$ to 40$\%$ over current state-of-the-art methods, and better qualitative results. Our promising results may indicate a paradigm rollback to the \textit{fully-explicit} and \textit{geometry-centric} algorithm design. \revise{In addition, we newly contribute a real-scanned 3D human dataset featuring more intricate geometric details.} We will make our code and data publicly available at \href{https://github.com/yztang4/HaP}{https://github.com/yztang4/HaP}.
\end{abstract}

\begin{IEEEkeywords}
    Human reconstruction, single-view image, point cloud, depth, SMPL.
\end{IEEEkeywords}

\section{Introduction}

\revise{Reconstructing the 3D geometric structures of human bodies from single-view 2D RGB images has been a fundamental and long-standing task, which lays the foundation for numerous downstream applications, such as immersive communication, video games, and virtual/augmented reality. However, it still remains highly challenging to produce accurate and detailed reconstruction results, due to the difficulty in inferring complete 3D human geometry information from heavily occluded regions in single-view image capture.}

In recent years, motivated by the efficiency and flexibility of deep implicit functions \cite{chen2019learning,mescheder2019occupancy,park2019deepsdf} to represent unconstrained-topology 3D shapes with fine-level geometric details, pixel-aligned implicit reconstruction pipelines \cite{saito2019pifu,saito2020pifuhd} have become the dominating processing paradigm due to their ability to recover high-fidelity geometry for clothed humans. However, these approaches typically overfit the limited training data due to the lack of human body priors, resulting in degenerate body structures and broken limbs when encountering novel poses or clothing styles. \revise{To overcome such limitations, the more recent trend \cite{zheng2021pamir,xiu2022icon} is to constrain the fully implicit learning mechanism by incorporating parametric 3D human body models \cite{SMPL:2015, SMPL-X:2019} as the explicit shape priors. However, although combining implicit representations with explicit geometric constraints is considered as a more promising exploration direction, it also poses additional challenges and higher requirements for the design of the specific combination scheme. The fusion of learned features or intermediate results inferred from input image pixels and the estimated parametric model typically leads to geometric over-smoothing and thus weakens surface details. More importantly, the topological inconsistency between the chosen parametric template model and the target human shape (especially with loose clothing or complex accessories) usually causes over-constraining effects. The latest effort \cite{xiu2022econ} jointly exploits the representation capability of implicit fields and robustness of explicit human body models, but the quality of the resulting reconstruction can still be constrained by unsatisfactory rectifications of parametric body models~\cite{SMPL-X:2019} and inaccurate predictions of normal maps.}

In practice, there are several aspects of key consideration when \revise{evaluating the performance} of the human reconstruction pipeline:
\begin{enumerate}
 \item flexibility to support unconstrained-topology modeling of human body shapes with arbitrary clothing;

 \item  generalizability to unseen data distributions (e.g., novel poses and clothes);

 \item  robustness to avoid non-human reconstruction results (e.g., unnatural poses, degenerate body structures, and broken limbs);

 \item  capability to represent and capture fine-level details of surface geometry in 3D space.
\end{enumerate}

In this paper, we seek to develop a learning-based single-view human reconstruction framework that simultaneously meets the above critical requirements, making it a more promising processing paradigm with greater potential compared with previous methods. Architecturally, we construct HaP, a fully explicit point-based modeling pipeline featuring direct point cloud estimation (from the \revise{2D image}), manipulation, generation, and refinement in the explicit 3D geometric space.

More specifically, given an input RGB image, our process begins with depth estimation to infer a partial 3D point cloud representing high-fidelity visible-view geometry. Meanwhile, we also estimate the SMPL model from the 2D image space to provide the missing human body information. 
Under the assumption that the estimated depth maps are generally accurate, the preceding depth-\revise{inferred} partial 3D points are further exploited for pose correction through a specialized SMPL rectification procedure to make the estimated SMPL model register well with the points \revise{in the 3D space}. 
Directly merging the 3D information of both the depth-\revise{inferred} partial points and the estimated SMPL model can coarsely form a complete 3D human point cloud with rich texture and wrinkle details; however, there are still apparent gaps between the partial points and the estimated SMPL model, which makes \revise{the combination of the partial points and the estimated SMPL model} unnatural and possibly problematic. To this end, we customize a diffusion-style
point cloud generation framework to learn the latent spatial distribution of realistic human bodies based on the coarse merging 3D human points, to generate 3D human bodies with clothing and pose information consistent with the depth-\revise{inferred} partial 3D points and the SMPL model. 
Moreover, we propose a refinement stage, including learning-based \revise{displacement} estimation and a simple yet effective depth replacement strategy, to further enhance the geometric quality. Finally, a high-quality mesh model can be derived from the point cloud using standard surface reconstruction methods, such as screened Poisson surface reconstruction \cite{kazhdan2013screened}. 

\begin{figure*}[t]
    \hsize=\textwidth 
    \centering
    \includegraphics[width=6.8in]{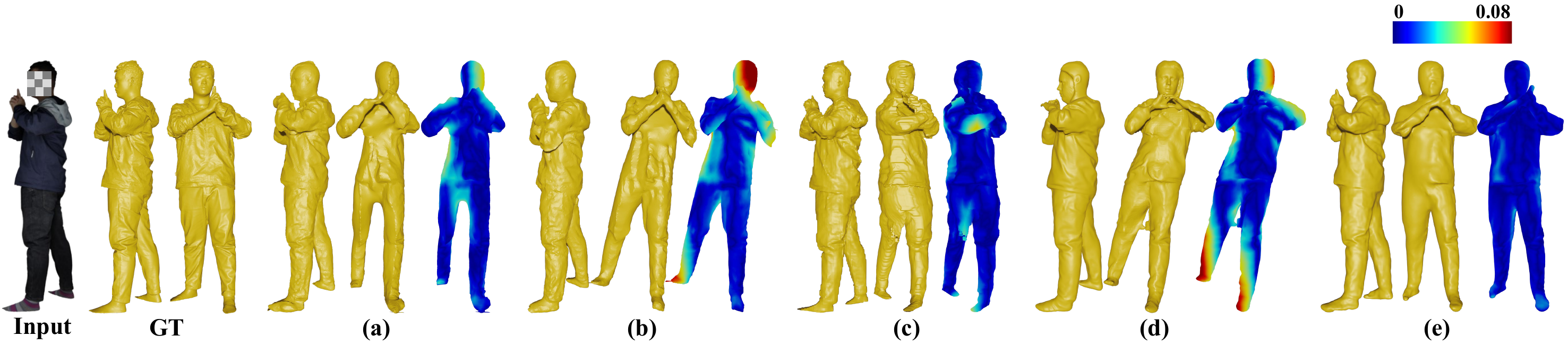}
    \caption{Visual comparisons of reconstructed human bodies and distance error maps by different methods.  (a) PIFu \cite{saito2019pifu}, (b) ICON \cite{xiu2022icon}, (c) IntegratedPIFu \cite{chan2022integratedpifu}, (d) ECON \cite{xiu2022econ}, (e) Proposed HaP. Our method can reconstruct clothing details and poses better than existing methods. \textcolor{red}{\faSearch} Zoom in for detailed geometry. }
    \label{fig:teaser}
\end{figure*}

\begin{figure*}[h]
    \centering
    \includegraphics[width=6.94in]{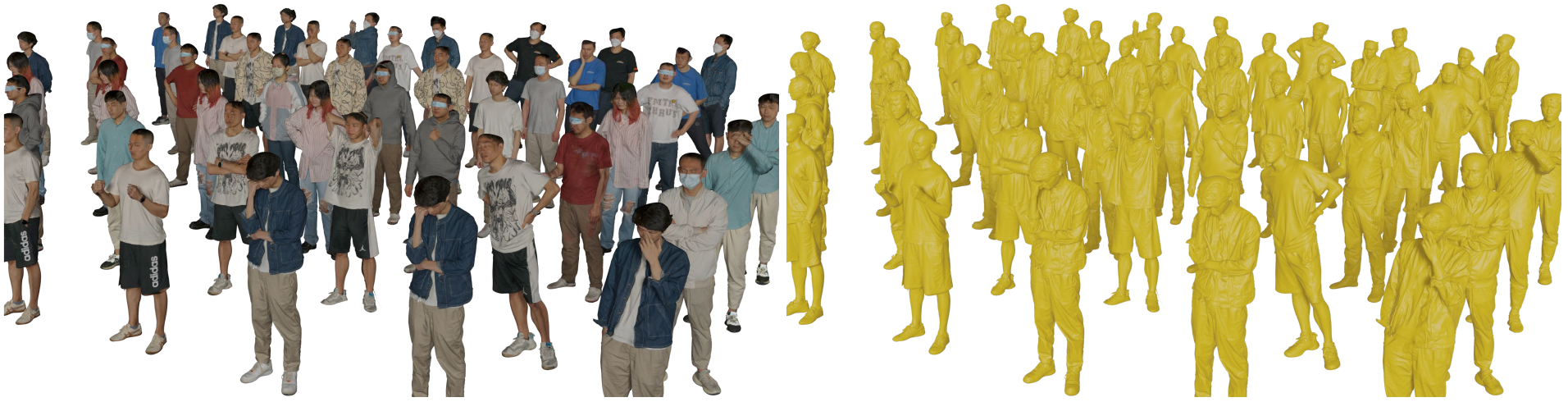}
    \caption{Illustration of the samples with texture and geometry details in our proposed dataset named CityUHuman. \textcolor{red}{\faSearch} Zoom in for detailed geometry. \revisesec{We also refer the readers to the \textbf{cityuhumanvideodemo.mp4} in the supplementary material.}}
    \label{CityUHumansamples}
\end{figure*}
With the design of the dedicated components, the proposed HaP framework comprehensively addresses the concerns mentioned above. Concretely,
\begin{enumerate}
    \item we utilize point clouds as the intermediate representation, which is known to be flexible \cite{qian2020pugeo,ren2023geoudf} for modeling arbitrary-topology geometric structures;

    \item the process of inferring coarse 3D geometry information from the \revise{2D image} consists of 2D depth estimation and SMPL estimation. The former, as a richly investigated and relatively mature task, shows satisfactory generalizability. The latter improves reconstruction robustness by injecting human body priors without destroying clothing cues mined from the specific input image; and

    \item it is more straightforward and effective to rectify poses and capture surface details when the point-based learning process is directly implemented in the original 3D geometric space.
\end{enumerate}

Extensive experiments demonstrate that our HaP outperforms state-of-the-art approaches both quantitatively and qualitatively (see Fig. \ref{fig:teaser}). 

In summary, the major contributions of this paper are as follows.

\begin{itemize}
    \item we propose a new pipeline, HaP, to explicitly generate human bodies in point clouds through a diffusion process conditioned on depth maps and SMPL models in 3D space; 
    \item we propose an effective SMPL rectification module to optimize SMPL models to accurate poses; 
    \item \revise{we contribute a new human scan dataset comprising fine-grained, detailed scans to facilitate future research endeavors, as illustrated in Fig. \ref{CityUHumansamples}.}
\end{itemize}

The remainder of this paper is organized as follows. Section \ref{sectionrelatedwork} provides a comprehensive review of the existing literature, including monocular depth estimation, point cloud generation, human pose and shape estimation and single-view human reconstruction. Section \ref{sectionpipeline} introduces our proposed HaP in detail. In Section \ref{sectionexp}, we conduct extensive experiments and ablation studies to \revise{evaluate} the effectiveness of HaP. Finally, Section \ref{sectionconclusion} concludes this paper.

\section{Related Work}
\label{sectionrelatedwork}
\revise{The HaP pipeline consists of three main modules: depth estimation from single-view RGB images, SMPL estimation and rectification using single-view RGB images, and conditional human point cloud generation. Based on these modules, we provide a concise overview of the related work in \textbf{Monocular Depth Estimation}, \textbf{SMPL Estimation and Rectification}, and \textbf{Point Cloud Representation}. Additionally, we briefly review other methods in \textbf{Clothed Human Reconstruction}.}

\subsection{Monocular Depth Estimation}

Predicting depth maps from single-view RGB images is a challenging \revise{task. Over} the years, various approaches~\cite{patni2024ecodepth,saxena2008make3d,eigen2014depth,lee2019big,patil2022p3depth,xie2023revealing} have been developed to tackle this problem. Make3D \cite{saxena2008make3d} divides the image into small homogeneous patches and uses a Markov random field to infer the plane parameters, which capture the location and orientation of each patch in 3D space. Eigen \textit{et al}.~\cite{eigen2014depth} proposed a pioneering deep learning-based method to learn an end-to-end mapping from RGB images to depth maps. BTS~\cite{lee2019big} designs novel local planar guidance layers for more effectively utilizing the densely encoded feature at multiple stages when decoding. P3Depth \cite{patil2022p3depth} iteratively and selectively leverages the information from coplanar pixels to improve the quality of the predicted depth. Xie \textit{et al}.~\cite{xie2023revealing} found that Masked image modeling (MIM) could also achieve state-of-the-art performance on depth estimation. 

Particularly, various methods have been proposed for estimating human depth maps from single-view images. Tang \textit{et al}. \cite{tang2019neural} employed a segmentation net and a skeleton net to generate human \revise{body parts and keypoints heatmaps}. These heatmaps were used to compute the base shape and detail shape via a depth net. Tan \textit{et al}. \cite{tan2020self} proposed a self-supervised method that utilizes the predicted SMPL models from videos. Similarly, HDNet~\cite{jafarian2021learning} predicts the corresponding dense pose of each image. Both Tan \textit{et al}. \cite{tan2020self} and Jafarian \textit{et al}. \cite{jafarian2021learning} warped body parts to different frames of a video and supervised the depth maps with a photo-consistency loss, as the depth map would not change much among different frames.

\revise{Building on the latest monocular depth estimation method~\cite{xie2023revealing}, we extract precise 3D information from the front view of the human body. This depth information acts as supervision for SMPL rectification and provides essential conditions for the diffusion model, significantly enhancing the accuracy of the overall reconstruction. By emphasizing depth as a cornerstone of the task, this approach achieves substantial improvements over previous methodologies.}

\revise{\subsection{SMPL Estimation and Rectification}}
\revise{The human body structure provides essential prior information for human body reconstruction.} Estimating 2D keypoints~\cite{alphapose, cao2017realtime} or human parsing~\cite{chen2023beyond, liang2015deep} can offer valuable insight into human structure. However, \revise{both 2D keypoints and human parsing} lack the spatial information necessary for precise 3D reconstruction. In contrast, predicting parametric models, such as SMPL(-X)~\cite{SMPL:2015}, from RGB images enables a more comprehensive understanding of human pose and shape, offering stronger prior knowledge for 3D human reconstruction.

Zhang \textit{et al}.\cite{zhang2021pymaf} proposed a regression-based approach that uses a feature pyramid and corrects the predicted parameters by aligning the SMPL mesh and image. PIXIE \cite{feng2021collaborative} introduces a moderator that merges the body, face, and hand features of experts according to their confidence weights. \revise{Kocabas \textit{et al}.\cite{kocabas2020vibe} proposed an adversarial learning framework to produce kinematical SMPL motion sequences in the absence of in-the-wild 3D ground truth.} METRO~\cite{lin2021end} attempts to model the interactions of \revise{vertex-vertex pairs} and \revise{vertex-joint connections}, and finally outputs human joint coordinates and SMPL vertices together. 

However, predicted SMPL(-X) \cite{SMPL:2015,SMPL-X:2019} models sometimes exhibit misalignment issues when observed in 3D space. \revise{ Xiu \textit{et al}.~\cite{xiu2022econ,xiu2022icon} utilized the rendering-based optimization method to rectify the SMPL poses. Zheng \textit{et al}. \cite{zheng2021pamir} optimized the SMPL parameters in the 3D implicit space. In this paper, we propose a novel SMPL rectification module that minimizes the distance between the visible \revise{part} of the SMPL model and the depth-\revise{inferred} partial point cloud in 3D space, and generate SMPL models that are more accurate and better-aligned with the given human body.}

\subsection{Point Cloud Representation}
\revise{Point clouds are an efficient and effective representation of 3D shapes, often serving as a precursor for generating other 3D formats, such as meshes and voxels. They have wide-ranging applications across various fields.} 

\textbf{Point Cloud Generation.} Generating point clouds can be broadly categorized into two groups: unconditioned and conditioned. RGAN \cite{achlioptas2018learning} is a classic unconditioned point cloud generation method, it uses a GAN structure with several MLPs to adversarially generate point clouds. WarpingGAN \cite{tang2022warpinggan} is a lightweight and efficient network, which warps several uniform 3D grids into point clouds with various resolutions. ShapeGF \cite{cai2020learning} first introduces the score-based network to unconditionally generate point clouds, it takes noise as input and aims to learn the implicit surface of 3D shapes. Conditioned generation tasks usually generate point clouds from RGB images \cite{fan2017point}, text information \cite{nichol2022point} or partial point clouds \cite{pan2021variational,zhou20213d,lyu2021conditional}. Fan \textit{et al}.~\cite{fan2017point} proposed a PointOutNet to predict multiple plausible point clouds when conditioned on RGB images. Point-E \cite{nichol2022point} first synthesizes a single-view RGB image using a text-to-image diffusion model, and then uses another diffusion model to generate a point cloud based on the generated image. The VRCNet \cite{pan2021variational} generates point clouds conditioned on partial point clouds, it considers the completion task as a relational reasoning problem and learns to reason about the spatial relationships between the points and complete the missing parts accordingly. PVD \cite{zhou20213d} and PDR \cite{lyu2021conditional} are diffusion-based networks, they also operate under conditioned scenarios, taking partial point clouds as conditions to complete the missing parts.

\revise{\textbf{Point-based Human Modeling.} Ma \textit{et al}.~\cite{ma2021scale} initially endeavored to represent clothed human body surface by learning point clouds, which can represent thin structures and open surfaces. POP \cite{ma2021power} proposes a novel dense point cloud representation that can model high-quality clothing shapes with various styles. Lin \textit{et al}.~\cite{lin2022learning} first learned implicit templates to represent coarse clothing topology and then generated point sets to enhance clothing deformations.}

\revise{ HaP also utilizes point clouds to represent the human body.  It reimagines human body reconstruction by reframing classic implicit approaches as an explicit point cloud generation task. Utilizing depth maps and SMPL models inferred from RGB images, HaP adopts a diffusion-based framework to produce detailed and complete human body point clouds, offering a fresh perspective on human body reconstruction.}

\begin{figure*}[t]
    \centering
    \includegraphics[width=7.in]{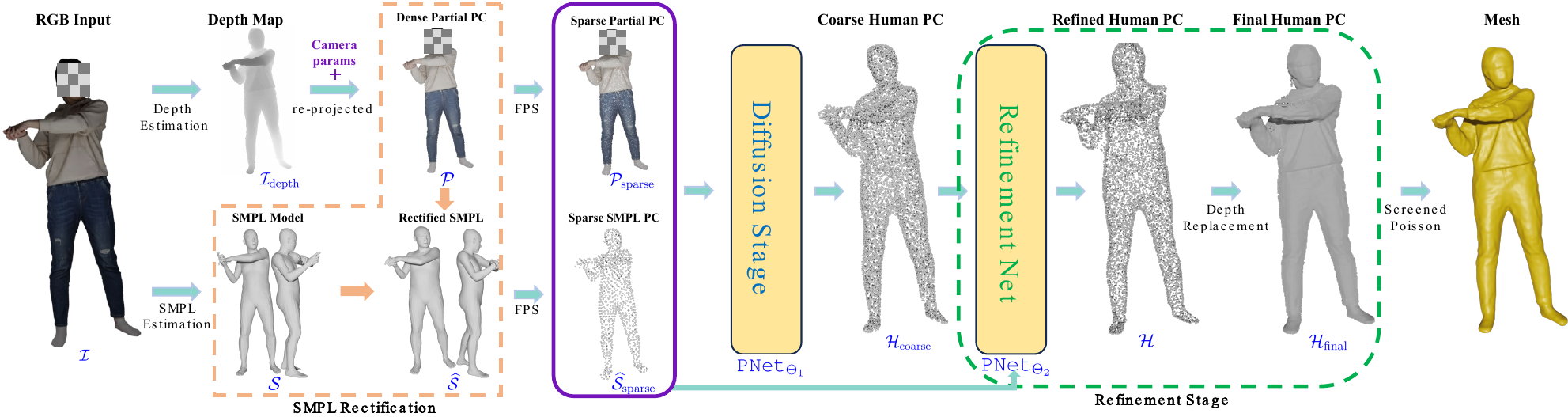}
    \caption{The pipeline of our framework HaP.  HaP first estimates \revise{ a depth map and an SMPL model} from the RGB input, which serve as conditions of a diffusion process to generate sparse human point clouds. At the refinement stage, we further propose a refinement network $\texttt{PNet}_{\Theta_2}(\cdot)$ and a depth replacement operation to enhance the quality of $\mathcal{H}_{\mathrm{coarse}}$, and finally reconstruct meshes from $\mathcal{H}_{\mathrm{final}}$ via screened Poisson \cite{kazhdan2013screened}. \revise{PC: Point Cloud. FPS: Farthest Point Sampling \cite{qi2017pointnet++}.}}
    \label{pipeline}
\end{figure*}

\subsection{Clothed Human Reconstruction}
\revise{\textbf{Implicit-based Approaches.} PIFu~\cite{saito2019pifu} represents a pioneering effort in integrating implicit functions for human reconstruction. Building upon this, PIFuHD~\cite{saito2020pifuhd} improves performance on high-resolution images through a multi-level architecture, trained with both front and back normal maps. Arch~\cite{huang2020arch} and Arch++\cite{he2021arch++} focus on learning implicit functions within canonical spaces, enabling the reconstruction of animatable human bodies. Additionally, Xiu \textit{et al}.\cite{xiu2022icon} and Zheng \textit{et al}.\cite{zheng2021pamir} incorporate SMPL models to provide prior knowledge of human body structure. Alldieck \textit{et al}.\cite{alldieck2022photorealistic} enhance visual fidelity by computing albedo and shading details, introducing an innovative rendering loss function. S3Fs~\cite{corona2023structured} enables the creation of relightable and animatable avatars. }

\revise{\textbf{Explicit-based Approaches.}  SiCloPe~\cite{natsume2019siclope} employs a generative adversarial network (GAN) conditioned on RGB images to synthesize consistent silhouettes and generate textured 3D meshes. More recently, Xiu \textit{et al}.\cite{xiu2022econ} introduced ECON, which simultaneously addresses both implicit and explicit reconstruction by generating 2.5D meshes from learned frontal and back normal maps. Tang \textit{et al}.\cite{tang2023high} designed a two-stage 3D convolutional network to explicitly learn the TSDF volumes of the human body.}

\revise{\textbf{Depth-guided Approaches.} Several methods~\cite{xie2023revealing,xiong2022pifu,jafarian2021learning,cao2022bilateral} have explored the role of depth information in 3D human body reconstruction. FAX~\cite{smith2019facsimile} converts single-view images into depth and albedo maps in 2D space to produce full 3D scans. Gabeur \textit{et al}.~\cite{gabeur2019moulding} trained a GAN to predict both front and back depth maps, converted them into point clouds, and used the screened Poisson method to reconstruct a human body. Similarly, Han \textit{et al}. \cite{2k2k} introduced a two-stage depth estimation network that learns front and back depth maps based on different body parts. Wang \textit{et al}.\cite{wang2020normalgan} trained an adversarial network to denoise the depth collected by RGB-D cameras and generate geometric details of the back view. Xiong \textit{et al}.\cite{xiong2022pifu} proposed a depth-guided self-supervised strategy to learn the implicit surface with signed distance function (SDF) values instead of occupancy values. Chan \textit{et al}.\cite{chan2022integratedpifu} designed a depth-oriented sampling scheme to learn high-fidelity geometric details. They achieved this by incorporating depth values as supplementary features at the pixel level, which served as the input for implicit functions.}

\revise{The proposed HaP introduces a novel explicit reconstruction framework by integrating depth information and the SMPL model. This approach utilizes a diffusion model to generate a complete human body point cloud within 3D space. Furthermore, the depth replacement technique enhances geometric details in the front view, bridging the gap between depth-guided and explicit reconstruction methods and improving performance over current approaches.}

\subsection{\revisesec{3D Generative Models}}
\revisesec{Significant progress has been made in advancing deep generative models for 3D content creation.
 The representative optimization-based approaches such as DreamFusion~\cite{poole2023dreamfusion} and Magic123~\cite{qian2023magic123}, leverage pre-trained 2D image diffusion models by score distillation, which often struggle with geometric inconsistency across views.  DreamCraft3D \cite{sun2023dreamcraft3d} introduces a hierarchical 3D generation framework that ensures geometric consistency and high-fidelity textures, incorporating bootstrapped score distillation to enhance appearance quality. Huang \textit{et al.} \cite{huang2024humannorm} proposed a text-to-3D model utilizing normal-adapted and depth-adapted diffusion models for detailed geometry generation, alongside a normal-aligned diffusion model to improve appearance realism. Li \textit{et al.} developed PSHuman \cite{li2024pshuman}, which employs a diffusion model to generate multi-view images and normal maps conditioned on an SMPL-X prior, followed by explicit human body surface reconstruction.}

\revisesec{Some methods aim to generate 3D contents directly in 3D space. Ren \textit{et al.} \cite{ren2024xcube} introduced a hierarchical voxel-diffusion generative model capable of producing large-scale 3D objects. Hunyuan3D \cite{zhao2025hunyuan3d} learns a latent representation of 3D objects and employs DiT \cite{peebles2023scalable} to train a generative model that supports 3D painting. Unlike the above-mentioned 3D generative methods, HaP learns a conditional diffusion model and manipulates point clouds entirely in 3D space rather than 2D image space, providing a more intuitive and direct learning and manipulation process for human body reconstruction.}

\section{Proposed Method} \label{sectionpipeline}

\subsection{Overview}

The overall processing pipeline of the proposed single-view 3D human reconstruction framework consists of two sub-tasks: (\textbf{1}) inferring and rectifying 3D geometric information from the input 2D RGB image; and    (\textbf{2}) generating a high-quality 3D human point cloud conditioned on the produced 3D geometric information, after which mesh reconstruction can be directly performed.

As illustrated in Fig. \ref{pipeline}, our HaP begins with two parallel branches for depth and SMPL estimation, which show \textbf{complementary} characteristics in terms of inferring 3D information of human geometry. Specifically, depth estimation is able to capture detailed geometric patterns of the front \revise{depth map $\mathcal{I}_\mathrm{depth}$} but requires additional efforts to recover invisible body parts; differently, the estimated SMPL \revise{$\mathcal{S}$} provides strong human body priors and avoids non-human structures (e.g., unnatural pose, degenerate body, broken limb) but suffers from fixed-topology and over-smoothed surface details. Thus, we are motivated to jointly exploit such two sources of 3D information. 
Particularly, we propose an optimization-based SMPL rectification module to align the initially estimated SMPL model \revise{$\mathcal{S}$} with the depth-\revise{inferred} partial 3D point cloud \revise{$\mathcal{P}$}. After unifying the geometric information of depth and rectified SMPL \revise{$\widehat{\mathcal{S}}$} into a 3D point cloud, which could still be problematic due to the inevitable gap between depth and SMPL cues, we use a conditional point diffusion model, which takes both the depth-\revise{inferred} sparse partial point cloud \revise{$\mathcal{P}_\mathrm{sparse}$} and SMPL-\revise{inferred} sparse point cloud  \revise{$\widehat{\mathcal{S}}_\mathrm{sparse}$} as inputs to produce a more accurate 3D point cloud of complete human body point cloud \revise{$\mathcal{H}_\mathrm{coarse}$}. In addition, \revise{a} refinement stage is explored to further enhance the geometric quality of \revise{$\mathcal{H}_\mathrm{coarse}$} and get \revise{$\mathcal{H}_\mathrm{final}$}. Finally, we apply a classic surface reconstruction algorithm \cite{kazhdan2013screened} to directly extract a mesh representation.

\subsection{Estimating 3D Information from Single 2D Images}

\revise{Given a single-view 2D RGB image denoted as $\mathcal{I}$, we aim to extract two 3D geometric information from it, i.e., the depth information $\mathcal{I}_\mathrm{depth}$ and the SMPL model $\mathcal{S}$.}

\subsubsection{Depth estimation} As a classic task in computer vision, monocular depth estimation \cite{ming2021deep} has been extensively investigated and achieved remarkable advancements, which provides a generic way of \revise{lifting 2D information} captured by RGB images to spatial structures in the 3D geometric space.

In our implementation, we employ the recent state-of-the-art monocular depth estimator MIM \cite{xie2023revealing} to predict the corresponding depth map $\mathcal{I}_\mathrm{depth}$ of $\mathcal{I}$. Then we project the depth value at each 2D-pixel coordinate $(u,v)$ to a 3D point $(x, y, z)$ according to the camera's parameters. \revise{(This work adopts the setting of PIFu \cite{saito2019pifu}, assuming a fixed camera model with parameters to convert depth maps into 3D partial point clouds, without predicting camera parameters for individual images.)} Besides, we also append the original RGB value of each pixel, \revise{producing} a colored 3D point denoted as $\textbf{p}=(x, y, z, r, g, b)$. Thus, \revise{by leveraging rembg \cite{githubGitHubDanielgatisrembg} (or methods like Segment Anything \cite{kirillov2023segment}) to obtain the valid mask of the human body, we are able to extract the useful pixels and subsequently eliminate the invalid ones}, consequently obtain a 3D partial human body point cloud denoted as $\mathcal{P}=\{\textbf{p}\}$.

\subsubsection{SMPL Estimation and Rectification}  
\begin{figure}[t]
    \centering
    \includegraphics[width=3.3in]{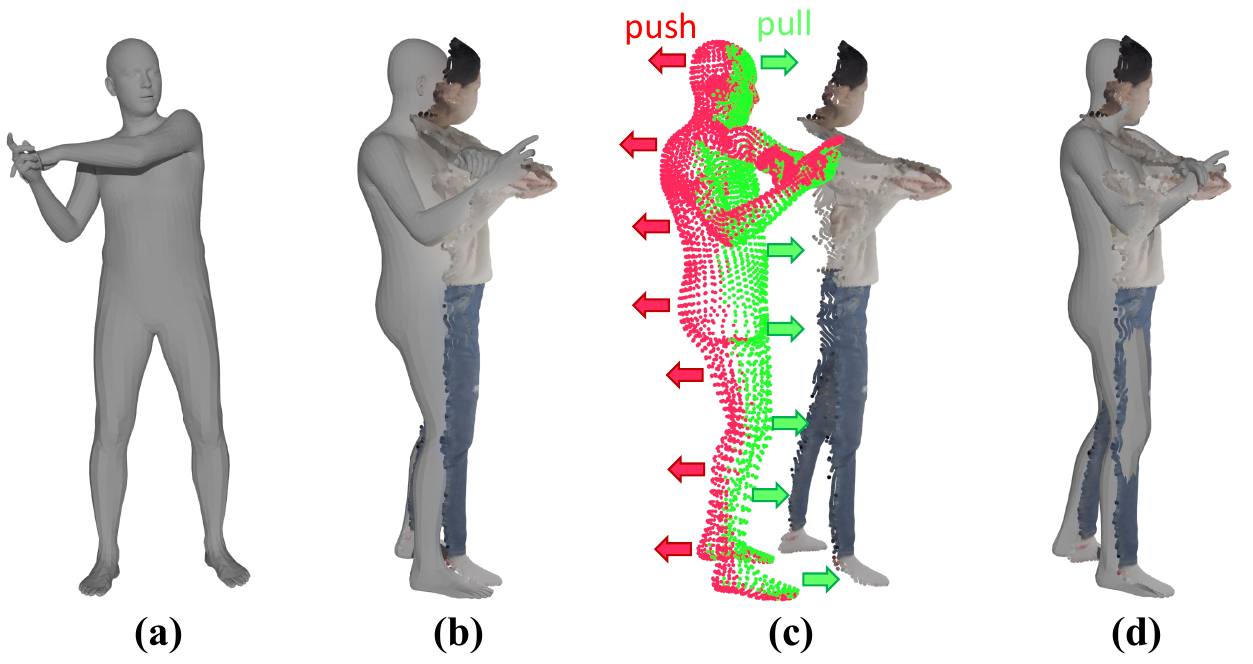}
    \caption{Illustration of the SMPL rectification process. (a) Initially estimated SMPL model. (b) Registration before rectification. (c) Rectification process. (d) Registration after rectification.  The directly estimated SMPL model is not well registered with the partial point cloud, and the situation is significantly relieved after the rectification process.}
    \label{smplrefineprocess}
\end{figure}

To provide explicit human shape priors and supplement the missing body parts in the partial depth-\revise{inferred} point cloud $\mathcal{P}$, we estimate from $\mathcal{I}$ an SMPL model $\mathcal{S}$ following \cite{zhang2021pymaf, feng2021collaborative}. However, as analyzed in previous works \cite{zheng2021pamir}, there would be inevitable misalignment between $\mathcal{S}$ and $\mathcal{P}$ caused by depth ambiguity, \revise{as shown in Fig. \ref{smplrefineprocess} (b)}, especially under the single-image setup, which can directly degrade the quality of the whole human reconstruction pipeline. To address this issue, assuming that depth estimation usually shows high accuracy and thus can deduce $\mathcal{P}$ with satisfactory quality\footnote{In some extreme cases where the estimated depth map shows poor quality, one can switch to reliability-aware depth estimation models (e.g., \cite{heo2018monocular,xia2020generating,hornauer2022gradient}) and adaptively determine the utilization of depth information.}, we propose a novel SMPL rectification module to further promote the alignment of $\mathcal{S}$ with $\mathcal{P}$ by updating the SMPL parameters, \revise{the process is illustrated in Fig. \ref{smplrefineprocess} (c)}. 

Let $\mathcal{S}_{v}=(\mathcal{V}_v,~\mathcal{F}_v)$ and $\overline{\mathcal{S}}_{v}=(\overline{\mathcal{V}}_v,~\overline{\mathcal{F}}_v)$ be the {\it visible} and the {\it invisible} \revise{part}s of $\mathcal{S}$ (i.e. $\mathcal{S}=\mathcal{S}_v\cup \overline{\mathcal{S}}_v$), respectively, where $\mathcal{V}_v$, $\overline{\mathcal{V}}_{v}$, $\mathcal{F}_v$, and $\overline{\mathcal{F}}_v$ stand for the vertex and face sets of the two \revise{part}s, and $\widehat{\mathcal{S}}$ the rectified SMPL model. Generally, this problem can be achieved through iterative rectifications to the shape parameters $\bm{\beta}$ and pose parameters $\bm{\theta}$ of $\mathcal{S}$ under the objective of minimizing the distance between $\mathcal{S}_v$ and $\mathcal{P}$. 

 Besides, to avoid any unwanted scenarios where the invisible part gets unnaturally close to $\mathcal{P}$, resulting in unnatural poses, we add a regularization term to \revise{maximize} the distance between $\mathcal{P}$ and $\overline{\mathcal{S}}_v$. We also regularize 
 shape parameters $\bm{\beta}$ due to their inherent zero-mean nature as established \cite{bogo2016keep}. Moreover, the optimization is subject to two constraints:  (\textbf{1}) the difference between the 2D keypoints of $\widehat{\mathcal{S}}$ and $\mathcal{S}$, preventing excessive variations in the pose of $\widehat{\mathcal{S}}$ on the \revise{2D image}, and (\textbf{2}) the silhouette difference between the mask of $\widehat{\mathcal{S}}$ (i.e., $\mathcal{M}_{\widehat{\mathcal{S}}}$) when projected onto the \revise{2D image} and the human mask from $\mathcal{I}$ (i.e., $\mathcal{M}_h$), 
 preserving the overall human shape of $\widehat{\mathcal{S}}$. In all, we explicitly formulate the problem as 

\begin{equation}
\begin{aligned}
\operatorname*{\textcolor{black}{min}}_{\bm{\textcolor{black}{\beta}}, \bm{\textcolor{black}{\theta}}} ~ \textcolor{black} {\mathcal{L}_\mathrm{r1}}\textcolor{black}{=} & \textcolor{black}{\lambda_1 \texttt{P2F}(\mathcal{P},~\mathcal{S}_v)+\lambda_2 \texttt{CD}(\mathcal{V}_v,\mathcal{P})} - \\&\textcolor{black}{\lambda_3 \texttt{P2F}(\mathcal{P}, \overline{\mathcal{S}}_v) }\textcolor{black}{+\lambda_4 \bm{\beta}+\lambda_5 \texttt{L}_1(\mathcal{K},\mathcal{K}_0) }, 
\end{aligned}
\label{equ:depth-smpl-align}
\end{equation}
where $\texttt{P2F}(\cdot,~\cdot)$, $\texttt{CD}(\cdot,~\cdot)$ and $\texttt{L}_1(\cdot,~\cdot)$ denote the point-to-face distance (P2F), Chamfer distance (CD) and L1 loss, respectively; $\mathcal{K}$ and $\mathcal{K}_{\texttt{0}}$ are the 2D keypoints of the SMPL at each iteration and $\mathcal{S}$, respectively; the non-negative hyperparameters $\lambda_1 \sim \lambda_5$ balance different regularization terms, \revise{and $\lambda_1=10$, $\lambda_2=3$, $\lambda_3=0.2$, $\lambda_4=0.1$,  $\lambda_5=0.1$}.  \revise{As demonstrated in Fig. \ref{smplrefineprocess} (d), $\widehat{\mathcal{S}}$ is better aligned with the partial point cloud $\widehat{\mathcal{P}}$ after the rectification.}

\begin{figure}[t]
    \centering
    \includegraphics[width=3.3in]{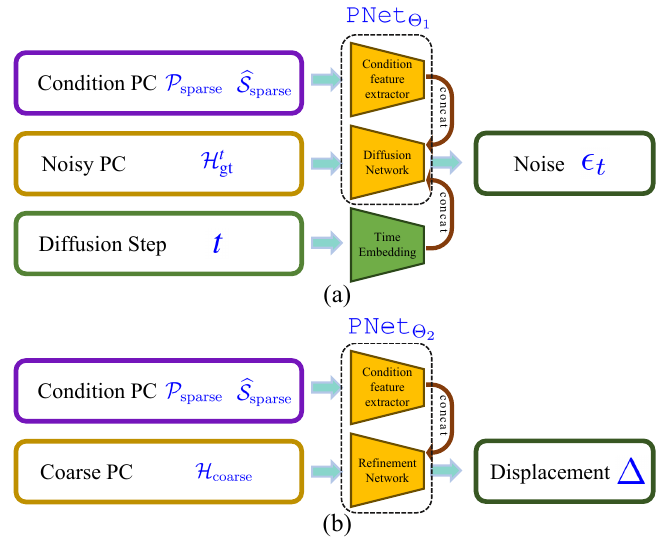}
    \caption{\revise{Illustration of the point cloud generation process. (a) Diffusion process. (b) Refinement process.} }
    \label{illustrationofdiffusion}
\end{figure}
\subsection{Diffusion-based Explicit Generation of Human Body}

The depth-\revise{inferred} partial human body point cloud $\mathcal{P}$ offers free-form yet detailed geometric structures such as wrinkles in clothes of the visible side, leaving the occluded side completely absent. 
\revise{Instead}, the rectified SMPL parametric human shape surface $\widehat{\mathcal{S}}$ is ensured to provide a complete body surface with reasonable pose and shape, while its fixed-topology causes essential difficulties in dealing with loose and complex clothing. These complementary characteristics motivate us to jointly exploit $\mathcal{P}$ and $\widehat{\mathcal{S}}$ for reconstructing high-fidelity 3D human geometry. Nevertheless, directly merging together the depth-\revise{inferred} and the SMPL-\revise{inferred} 3D geometric information can be problematic, since there usually exists a non-negligible gap between $\mathcal{P}$ and $\widehat{\mathcal{S}}$.

\revise{To achieve this, we propose training a conditional denoising diffusion probabilistic model (DDPM) to generate high-quality 3D human body point clouds. The model aims to naturally merge $\mathcal{P}$ and $\widehat{\mathcal{S}}$, eliminating any gaps. We sample sparse point clouds from $\mathcal{P}$ and $\widehat{\mathcal{S}}$ using farthest point sampling (FPS) \cite{qi2017pointnet++}, denoted as $\mathcal{P}_\mathrm{sparse}$ and $\widehat{\mathcal{S}}_\mathrm{sparse}$, respectively. The backbone of the diffusion model is a two-branch PointNet++ architecture \cite{lyu2021conditional, qi2017pointnet++}, parameterized by $\Theta_1$, denoted as $\texttt{PNet}{\Theta_1}(\cdot)$. As shown in Fig. \ref{illustrationofdiffusion} (a), a condition feature extractor extracts features from the sparse condition point clouds $\mathcal{P}_\mathrm{sparse}$ and $\widehat{\mathcal{S}}_\mathrm{sparse}$, while a time embedding layer extracts features of the diffusion step $t$. The features from the condition point clouds and the time step are then concatenated and used as input to the diffusion network. The diffusion network takes the noisy point cloud $\mathcal{H}_\mathrm{gt}^t$ and outputs the Gaussian noise $\epsilon_t$ at this diffusion step.}

\revise{In our experiments, we found that although the diffusion stage was implemented, it did not completely remove all noise, resulting in overly rough geometric information that was not suitable for direct reconstruction. To address this, we introduce a refinement stage after the diffusion process to improve the quality of the generated point cloud by eliminating the remaining noise. As illustrated in Fig. \ref{illustrationofdiffusion} (b), we use a refinement network, parameterized by $\Theta_2$, denoted as $\texttt{PNet}{\Theta_2}(\cdot)$, to refine the generated human point clouds $\mathcal{H}_\mathrm{coarse}$ by learning displacement $\Delta$ for each point. Additionally, we apply a depth replacement strategy to densify the generated point cloud.}

\subsubsection{Diffusion Stage}

At this stage, we train a conditional diffusion model to generate a human point cloud. Due to constraints imposed by limited GPU memory and computational time, we configure $\texttt{PNet}_{\Theta_1}(\cdot)$ to generate a relatively sparse $\mathcal{H}_\mathrm{coarse}$ with 10,000 points. During the training process, we use the farthest point sampling operation to sample points from $\mathcal{P}$ and $\widehat{\mathcal{S}}$ respectively, which are concatenated as conditions for the diffusion model. A conditional DDPM consists of two Markov chains in opposite directions, i.e., the forward process and the reverse process, both processes have the same sampling step length $T$. During the forward process, the Gaussian noise is \revise{added to} $\mathcal{H}_\mathrm{gt}$ at each sampling step $t$, and when $T$ is large enough, the output of the forward process should have \revise{a distribution similar to the Gaussian distribution} $\mathcal{N}(\boldsymbol{0},\boldsymbol{I})$, the forward process is from the initial step $\mathcal{H}_\mathrm{gt}^0$ to the last step $\mathcal{H}_\mathrm{gt}^T$. Note that the condition $(\mathcal{P},\widehat{\mathcal{S}})$ is not included in the forward process.

The sampling process of the forward direction is iterative, however, we cannot train the DDPM at each time step due to the significant computational cost. According to \cite{ho2020denoising}, we can achieve arbitrary step $\mathcal{H}_\mathrm{gt}^t$ by defining $1-\gamma_t=\alpha_t$, $\bar{\alpha_t} = \prod_{i=1}^t \alpha_i$, where $\gamma$ is a pre-defined hyperparameter. Thus we formulate $q(\mathcal{H}_\mathrm{gt}^t|\mathcal{H}_\mathrm{gt}^0)=\mathcal{N}(\mathcal{H}_\mathrm{gt}^t;\sqrt{\bar{\alpha_t}}\mathcal{H}_\mathrm{gt}^0,(1-\bar{\alpha_t})\boldsymbol{I})$, and $\mathcal{H}_\mathrm{gt}^t$ can be sampled through:
\begin{equation}
        \mathcal{H}_\mathrm{gt}^t=\sqrt{\bar{\alpha_t}}\mathcal{H}_\mathrm{gt}^0+\sqrt{1-\bar{\alpha_t}}\boldsymbol{\epsilon}, 
\end{equation}
where $\boldsymbol{\epsilon}$ is Gaussian noise. With a parameterization technique proposed in \cite{ho2020denoising}, we can train $\Theta_1$ by minimizing:
\begin{equation}
\begin{aligned}
    \mathcal{L}_\mathrm{ddpm} = & \mathbb{E}_{t,\mathcal{H}_\mathrm{gt}^t,\epsilon}\| \epsilon-\texttt{PNet}_{\Theta_1}(\mathcal{H}_\mathrm{gt}^t,\mathcal{P},\widehat{\mathcal{S}},t)\|^2
\end{aligned}
\end{equation}
During the training process, both $\mathcal{P}$ and $\widehat{\mathcal{S}}$ are included as conditions at each time step to provide supervision on clothing information and human body information. 
At the inference process, we generate $\mathcal{H}_\mathrm{coarse}$ following the reverse process, i.e., progressively sampling $\mathcal{H}_\mathrm{gt}^{t-1}$ from $\mathcal{H}_\mathrm{gt}^t$ where $t$ \revise{starts} from $T$. The generated $\mathcal{H}_\mathrm{coarse}$ possesses the same clothing information as the provided $\mathcal{P}$, and its pose remains consistent with the pose provided by $\widehat{\mathcal{S}}$. \\

\subsubsection{Refinement Stage} 

During inference, we observe that it is typically impossible for $\texttt{PNet}_{\Theta_1}(\cdot)$ to eliminate all the noise in $\mathcal{H}_\mathrm{coarse}$. Consequently, we can only recover a coarse surface from $\mathcal{H}_\mathrm{coarse}$. To overcome this problem, we train $\texttt{PNet}_{\Theta_2}(\cdot)$ parameterized with $\Theta_2$ to learn a displacement for each point in $\mathcal{H}_\mathrm{coarse}$, i.e., $\Delta=\texttt{PNet}_{\Theta_2}(\mathcal{H}_\mathrm{coarse},\mathcal{P},\widehat{\mathcal{S}})
$, \revise{to further remove the remaining noise in} $\mathcal{H}_\mathrm{coarse}$, \revise{which enhances the detailed geometric information}. We obtain the human point cloud $\mathcal{H}$ with $\mathcal{H} = \mathcal{H}_\mathrm{coarse}+\Delta$. We formulate this problem as:

\begin{equation}
\min_{\Theta_2} \texttt{D}(\mathcal{H},\mathcal{H}_\mathrm{gt}) + \alpha\texttt{R}_\mathrm{smooth}(\Delta),
\end{equation}
where $\texttt{D}(\cdot, \cdot)$ stands for a distance metric realized by \cite{ren2024measuring}, 
$\texttt{R}_\mathrm{smooth}(\cdot)$ is a spatial smoothness regularization term  written  as:
\begin{equation}
\texttt{R}_\mathrm{smooth}(\Delta) = \frac{1}{3N_{\mathrm{src}}K_s} \sum\limits_{x \in \mathcal{H}_\mathrm{coarse}} \sum\limits_{x^{\prime} \in \mathcal{N}(x)} \|\delta_x-\delta_{x^{\prime}}\|_2^2
,
\end{equation}
where \revise{ $N_\mathrm{src}$ is the number of points contained in $\mathcal{H}_\mathrm{coarse}$, $\mathcal{N}(\cdot)$ returns $K_s$ nearest neighbor points}, and $\delta_x\in \Delta$ stands for the displacement of a typical point.

\begin{figure}[t]
    \centering
    \includegraphics[width=3.5in]{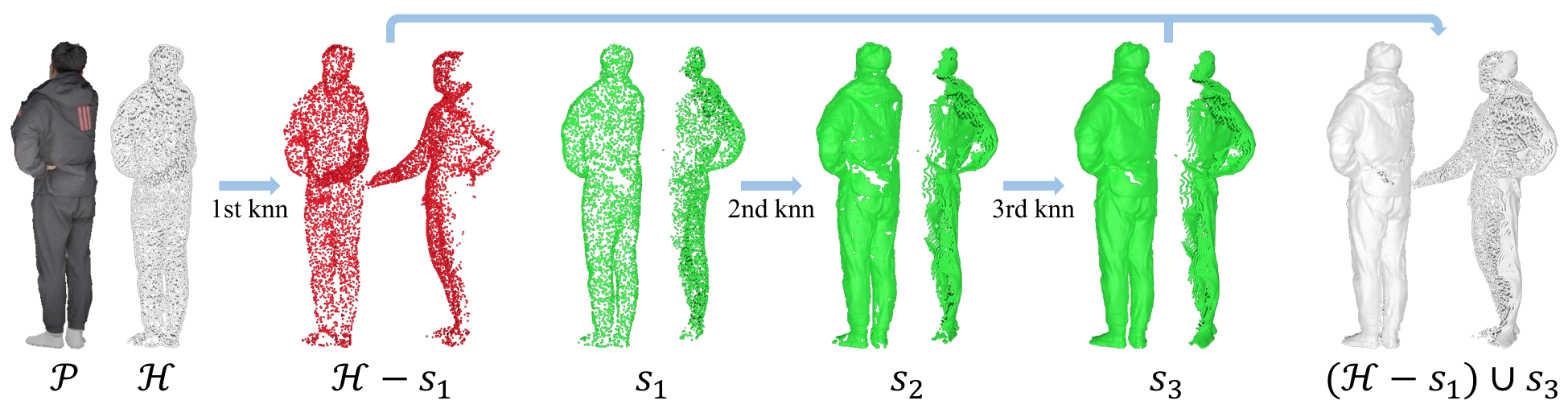}
    \caption{Illustration of the pipeline of the proposed depth replacement strategy. We first find \revise{dense partial point cloud} $\mathcal{P}$'s nearest points within \revise{refined human point cloud} $\mathcal{H}$ as $s_1$, and then we iteratively find $s_1$'s nearest points within $\mathcal{P}$, starting with $s_1$, then $s_2$, finally we achieve the \revise{final partial point cloud}  $\mathcal{H}_\mathrm{final}$, which is formed by combining \revise{occluded point cloud} $\mathcal{H}-s_1$ and \revise{visible point cloud} $s_3$.}
    \label{depthpipeline}
\end{figure}

In addition, in the preceding diffusion stage, we sample a relatively sparse set of 3D points as inputs to the DDPM for saving memory and computational costs, which determines the sparsity of $\mathcal{H}_\mathrm{coarse}$, as likewise $\mathcal{H}$. To produce a dense point cloud with enhanced geometric details, we specialize a simple yet effective depth replacement strategy that fully leverages the density of the preceding depth-\revise{inferred} \revise{dense partial point cloud} $\mathcal{P}$. As illustrated in Fig. \ref{depthpipeline}, our strategy commences by employing the ball query operation \cite{qi2017pointnet++} to locate the nearest points of $\mathcal{P}$ within \revise{refined human point cloud} $\mathcal{H}$; these nearest points are subsequently removed and denoted as $s_1$. Following this, we identify the nearest points of the \revise{occluded point cloud} $\mathcal{H}-s_1$ within $\mathcal{P}$ and designate them as $s_2$. Finally, we determine the nearest points to $s_2$ within $\mathcal{P}$ and label them as \revise{visible point cloud} $s_3$. The \revise{final human point cloud}, denoted as $\mathcal{H}_\mathrm{final}$, is formed by combining \revise{occluded point cloud} $\mathcal{H}-s_1$ and \revise{visible point cloud} $s_3$. 

After obtaining the \revise{final human point cloud} $\mathcal{H}_\mathrm{final}$, we directly adopt Screened Poisson \cite{kazhdan2013screened} for surface extraction.

\section{Experiments}
\label{sectionexp}
\subsection{Data Preparation}

\begin{figure}[t]
    \centering
    \includegraphics[width=3.2in]{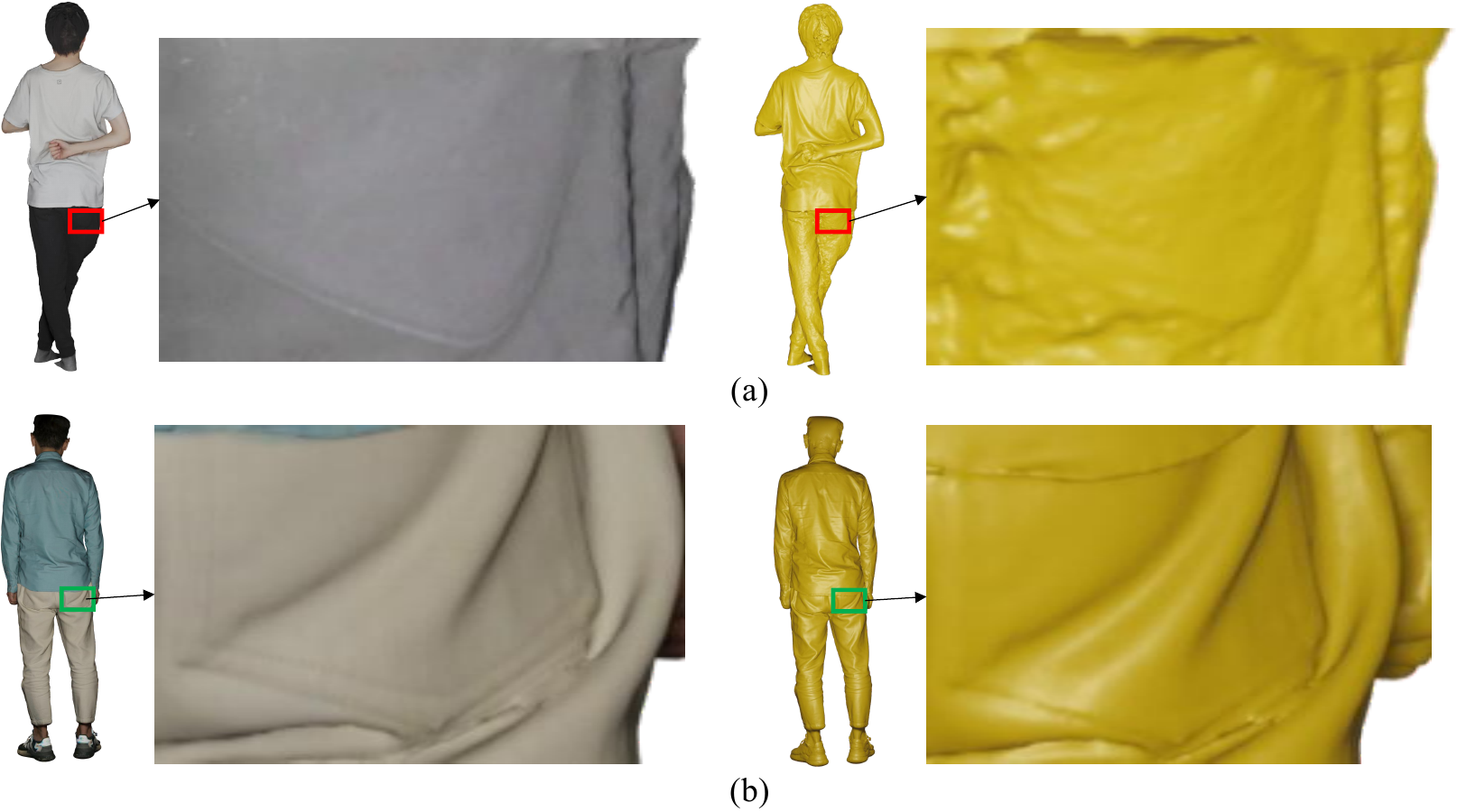}
    \caption{The visual comparison between (a) Thuman2 \cite{su2022deepcloth,tao2021function4d} and (b) CityUHuman. \textcolor{red}{\faSearch} Zoom in for detailed geometry.}
    \label{CityUHumanvsthuman}
\end{figure}

One of the primary development bottlenecks in the research area of single-view human reconstruction is the issue of \textit{data insufficiency}. 
To address this issue, we have created a new dataset, referred to as \textbf{CityUHuman}\footnote{\href{https://github.com/yztang4/HaP}{https://github.com/yztang4/HaP}}, which expands the number of available human body scans. We will make this dataset publicly available for research purposes \revise{so that it serves} as a valuable supplement for assessing and benchmarking the performance of single-view human reconstruction methods. \revise{ In comparison to the Thuman dataset \cite{su2022deepcloth,tao2021function4d}, the human models in the CityUHuman dataset exhibit smoother and more accurate surface details. As shown in Fig. \ref{CityUHumanvsthuman}, the Thuman2 scan displays a disrupted geometry at the pocket area due to its rough surface, whereas the CityUHuman scan preserves the finer geometric details more accurately.}

Specifically, we obtained 3D human body scans using a high-precision, portable 3D scanner named Artec Eva\footnote{https://www.artec3d.com/portable-3d-scanners/artec-eva}. The complete collection of each scan comprises over 10 million points. We  \revise{invited in total twelve volunteers and each volunteer was collected 3 to 11 scans}. Adhering to ethical principles and respecting personal information ownership, we obtained consent from each volunteer prior to the data collection process. Volunteers were informed that their 3D-scanned data would be exclusively available for non-commercial research applications. In an effort to protect the privacy of the volunteers, we kindly request users to blur the faces in any materials intended for publication, such as papers, videos, and posters.  Example scans with different clothes and poses from our dataset are illustrated in Fig. \ref{CityUHumansamples} to showcase its quality and potential for use in research.   

\subsubsection{Training data} In our experiments, we uniformly adopted Thuman2.0 \cite{tao2021function4d} for the training of our proposed HaP learning architecture. On the one hand, we extracted from each model 10000 3D points as the ground-truth point cloud. On the other hand, we employed the Blender software for rendering ground-truth depth maps and photo-realistic images with an interval of 10 degrees. Thus, for each RGB image serving as the input, we have three aspects of supervision signals, i.e., a depth map, an SMPL-\revise{inferred} point cloud, and a human point cloud. The total amount of the resulting training samples is 18000.

In particular, for making fair comparisons with IntegratedPIFu \cite{chan2022integratedpifu}, we further created a subset from our training samples to train our HaP by preserving the same scan indices used in the training process of IntegratedPIFu, while evenly selecting 10 views from each scan. 

\subsubsection{Testing data} We generated testing data from Thuman3.0 \cite{su2022deepcloth} and our collected CityUHuman datasets to evaluate the performances of different approaches. We rendered each scan using 8 views spanning every 45 degrees in the yaw axis, and then randomly selected 459 RGB images from Thuman3.0 and 40 RGB images from CityUHuman. In addition, we also collected several real images from the Internet and inferred the corresponding human bodies \revise{for a purely qualitative evaluation.}

\begin{figure*}[t]
    \centering
    \setlength{\abovecaptionskip}{0.1cm}
    \includegraphics[width=6.8in]{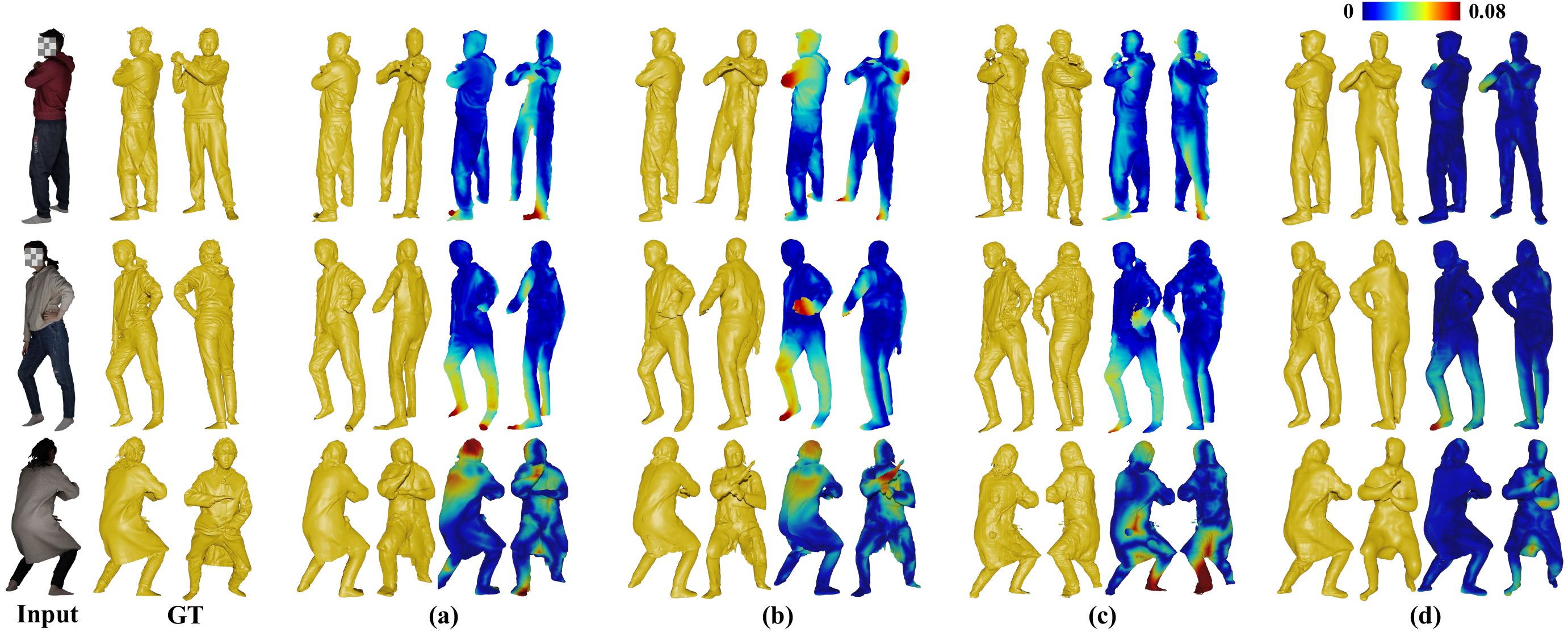}
    \caption{Visual comparisons of our HaP with implicit-based methods. The distance error maps of each reconstruction result are shown. (a) PIFu \cite{saito2019pifu}, (b) ICON \cite{xiu2022icon}, (c) IntegratedPIFu \cite{chan2022integratedpifu}, (d) our HaP. \textcolor{red}{\faSearch} Zoom in for detailed geometry.}
    \label{errormap}
\end{figure*}

\begin{figure*}[t]
    \centering
    \setlength{\abovecaptionskip}{0.1cm}
    \includegraphics[width=7in]{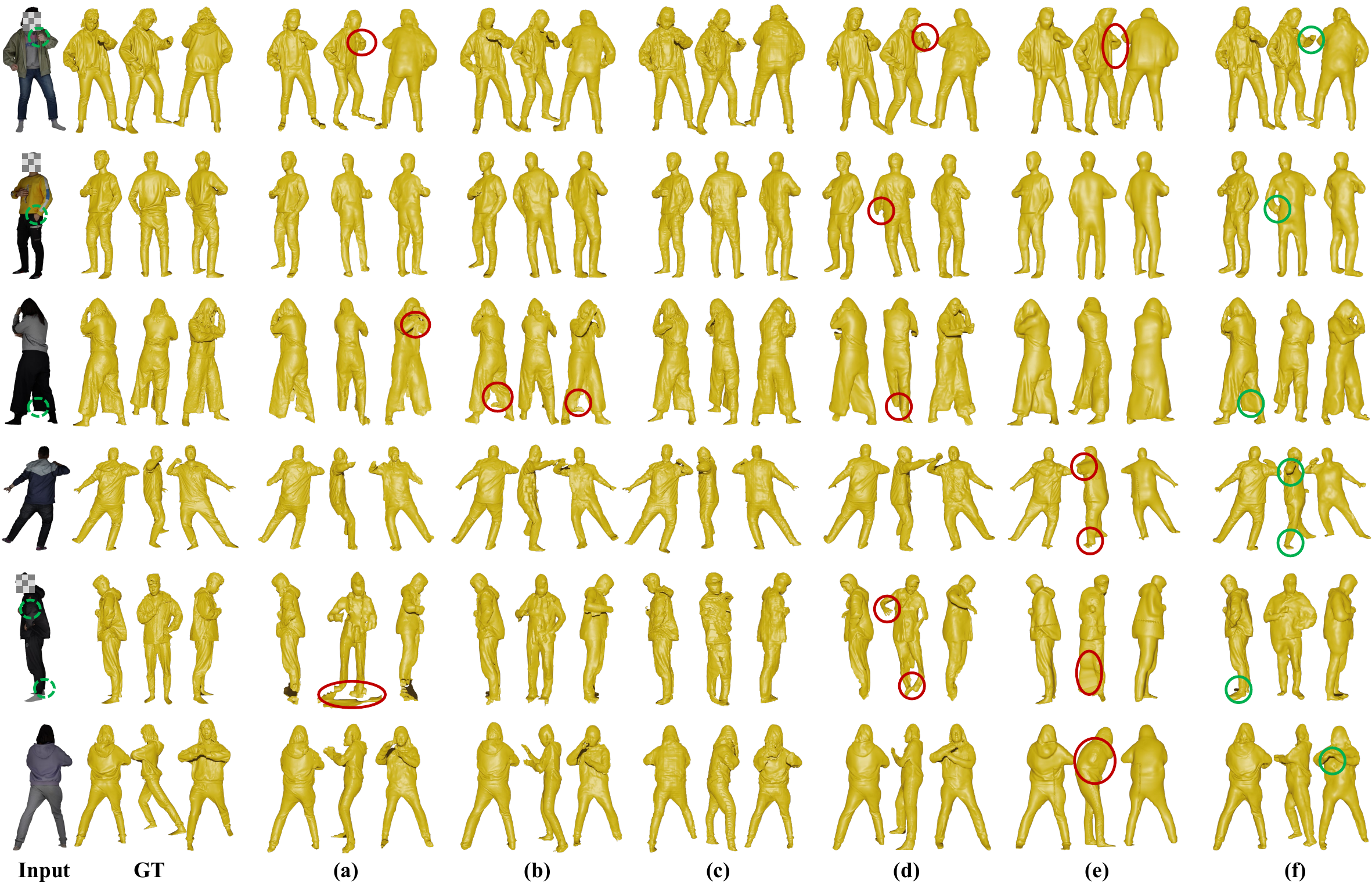}
    \caption{\revise{V}isual comparisons of our HaP with state-of-the-art methods. (a) PIFu \cite{saito2019pifu}, (b) ICON \cite{xiu2022icon}, (c) IntegratedPIFu \cite{chan2022integratedpifu}, (d) ECON \cite{xiu2022econ} (e) 2K2K \cite{2k2k} (f) our HaP.  \textcolor{red}{\faSearch} Zoom in for detailed geometry.}
    \label{compare1}
\end{figure*}

\begin{figure*}[t]
    \centering
    \setlength{\abovecaptionskip}{0.1cm}
    \includegraphics[width=7in]{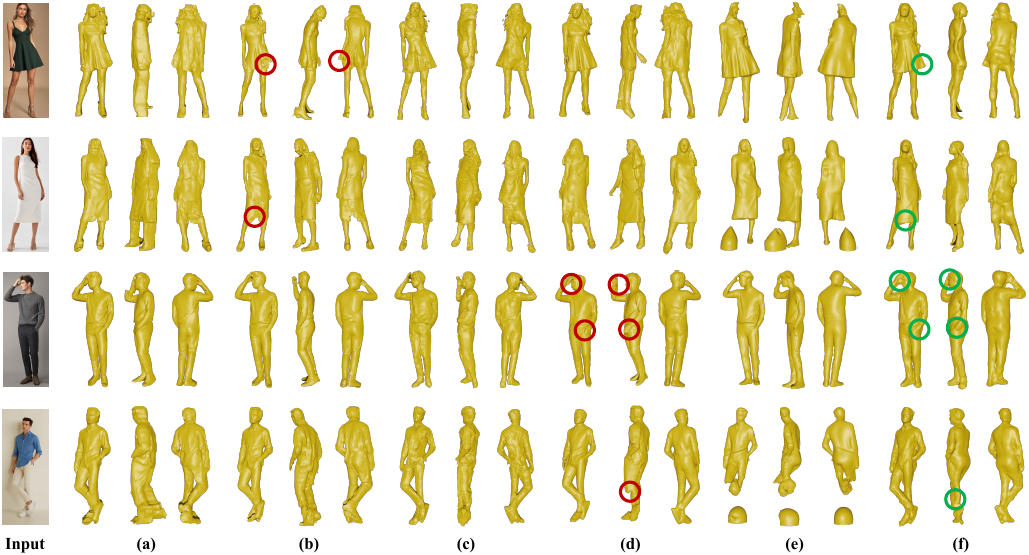}
    \caption{\revise{Visual comparisons of our HaP with state-of-the-art methods on the wild images. (a) PIFu \cite{saito2019pifu}, (b) ICON \cite{xiu2022icon}, (c) IntegratedPIFu \cite{chan2022integratedpifu}, (d) ECON \cite{xiu2022econ} (e) 2K2K \cite{2k2k} (f) our HaP. \textcolor{red}{\faSearch} Zoom in for detailed geometry.}}
    \label{compareinthwildoniconecon}
\end{figure*}

\begin{figure*}[t]
    \centering
    \includegraphics[width=6.5in]{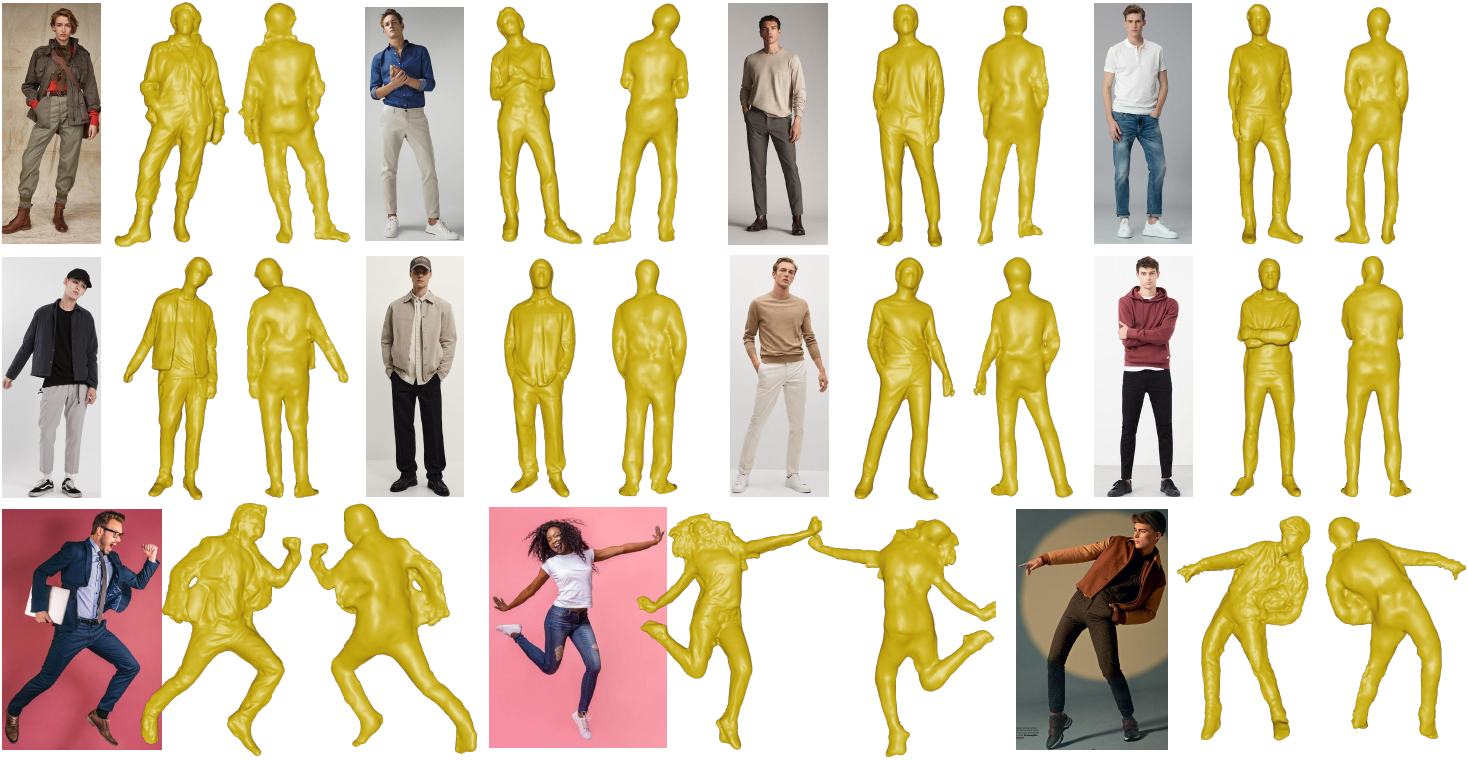}
    \caption{\revise{More results of our HaP applied to images captured in reality. We provide two views for each image. \textcolor{red}{\faSearch} Zoom in for detailed geometry.}}
    \label{wild}
\end{figure*}

\subsection{Implementation Details}

The employed depth estimation method MIM takes RGB images with a resolution of $512 \times 512$ as inputs. During the training process of MIM, we employed the \verb|swin-v2-base| architecture \cite{liu2021swin} as the backbone. The learning rate was set to $0.3\times 10^{-4}$, the batch size was set to 8, and the number of epochs was set to 30. The remaining parameters were kept the same as those in the original MIM. The training was performed on four RTX 3090 Ti GPUs.

During the training process of $\texttt{PNet}$,
since the SMPL models lack texture information, we set their color values to 0, and there was no color information present in the ground truth human point clouds ($\mathcal{H}_\mathrm{gt}$). At the refinement stage, we generated 18000 coarse human point cloud samples $\mathcal{H}_\mathrm{coarse}$ and randomly selected 1600 of these to train the refinement network to save time. Both the diffusion stage and the refinement stage had a batch size and epoch of 8 and 300, respectively. The remaining parameters were kept the same as in the original $\texttt{PNet}$ \cite{lyu2021conditional}. 

When rectifying the predicted SMPL models, we utilized the SGD optimizer to perform error backpropagation directly on the SMPL parameters $\bm{\beta},~\bm{\theta}$ for a total of 2000 iterations. The learning rate for the SGD was set to 0.03 and the momentum was set to 0.9.

We used FPS to uniformly sample 8192 points for $\mathcal{P}$, and 1024 and 2048 points for $\widehat{\mathcal{S}}$ at the diffusion stage and the refinement stage, respectively. We set the value of $k$ in the depth replacement module to 30 and the depth of Screened Poisson to 8.

\begin{figure}[t]
    \centering
    \setlength{\abovecaptionskip}{0.1cm}
    \includegraphics[width=3.5in]{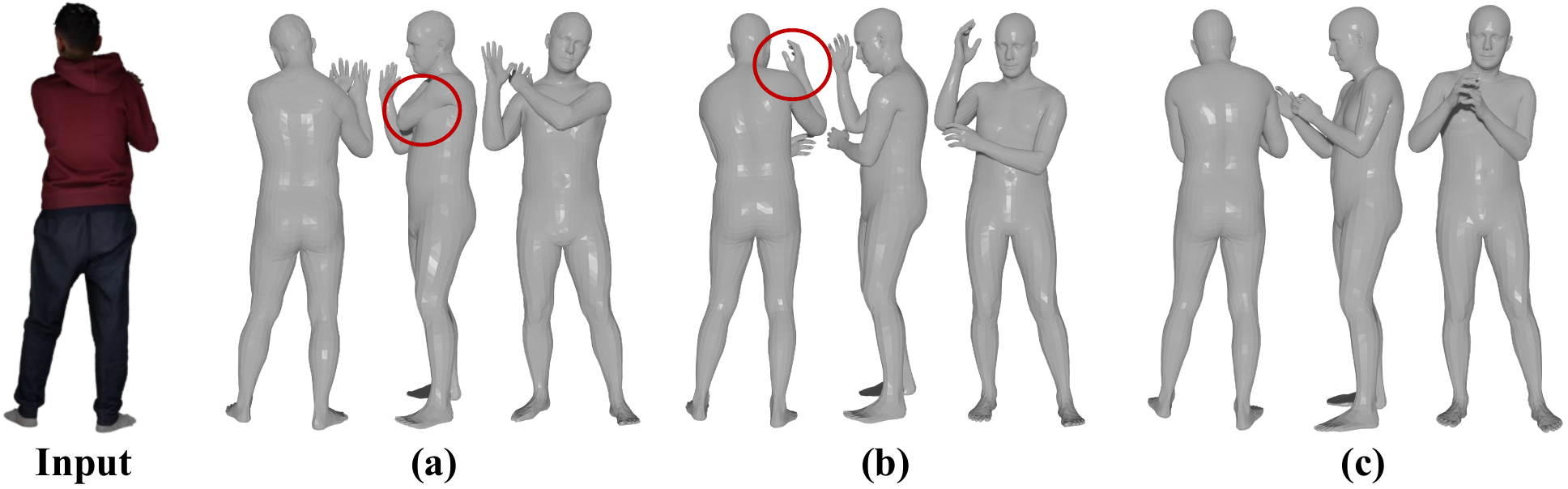}
    \caption{Comparisons of the results refined (a) with only $\texttt{P2F}$ loss, (b) without keypoint regularization and (c) with all regularizations.}
    \label{biandehebubiande}
\end{figure}

\begin{table*}[t]
    \centering
    \renewcommand\arraystretch{1.25}
    \setlength{\tabcolsep}{15.0pt}
    \caption{Quantitative comparisons of different methods on two datasets. \revise{ \textcolor{blue}{``$\downarrow$"} indicates that the lower values are better.} }
     \begin{tabular}{l|c|c|c|c|c|c} 
        \toprule[1.0pt]
        \multicolumn{1}{c|}{\multirow{2}[0]{*}{\diagbox{Method}{Metric}}} & \multicolumn{3}{c|}{Thuman3.0} & \multicolumn{3}{c}{CityUHuman} \\
        \cline{2-7}
        & \multicolumn{1}{c}{CD \textcolor{blue}{$\downarrow$}} & \multicolumn{1}{c}{P2F\textcolor{blue}{$\downarrow$}} & \multicolumn{1}{c|}{Normal\textcolor{blue}{$\downarrow$}} & \multicolumn{1}{c}{CD\textcolor{blue}{$\downarrow$}} & \multicolumn{1}{c}{P2F\textcolor{blue}{$\downarrow$}} & \multicolumn{1}{c}{Normal\textcolor{blue}{$\downarrow$}} \\
        \hline
        \hline
        PIFu  \cite{saito2019pifu}& 1.323   & 1.461  & 2.742  & 1.515  & 1.745  & 2.907 \\ 
        ICON   \cite{xiu2022icon} & 1.306   & 1.395 & 2.731  & 1.315   & 1.333 & 2.813 \\ 
        ECON  \cite{xiu2022econ} & 1.467   & 1.524   & 2.877 & 1.433   & 1.506   & 2.794 \\ 
        Our HaP & \textbf{0.860} & \textbf{0.925}   & \textbf{2.170} & \textbf{1.039}  & \textbf{1.173} & \textbf{2.371} \\ 
        \hline
        \color{gray}{IntegratedPIFu}   \cite{chan2022integratedpifu} & \color{gray}{1.148} & \color{gray}{1.182}  & \color{gray}{2.262}  & \color{gray}{1.140} & \color{gray}{1.187}  & \color{gray}{2.306}\\ 
        \color{gray}{Our HaP-I}       & \color{gray}{1.021} & \color{gray}{1.065} & \color{gray} {2.468} & \color{gray}{1.154} & \color{gray}{1.194}  & \color{gray}{2.523} \\
        \hline
        \color{gray}{2K2K}   \cite{2k2k} & \color{gray}{1.405} & \color{gray}{1.384}  & \color{gray}{8.184}  &\color{gray}{1.258} & \color{gray}{1.304}  & \color{gray}{8.052} \\ 
        \color{gray}{Our HaP-T} & \color{gray}{0.827} & \color{gray}{0.887} & \color{gray} {2.140} &  \color{gray}{1.043} & \color{gray}{1.174}  & \color{gray}{2.371}\\
        \bottomrule[1.0pt]
    \end{tabular}
    \label{sotaresults}
\end{table*}
\begin{table*}[t]
    \centering
    \renewcommand\arraystretch{1.25}
    \setlength{\tabcolsep}{15.0pt}
    \caption{Quantitative comparisons of visible and occluded sides on Thuman3.0.
     ``$\downarrow$" indicates that the lower values are better.
    }
    \begin{tabular}{l|c|c|c|c|c|c} 
        \toprule[1.0pt]
        \multicolumn{1}{c|}{\multirow{2}[0]{*}{\diagbox{Method}{Metric}}} & \multicolumn{3}{c|}{\textbf{Visible Side}} & \multicolumn{3}{c}{\textbf{Occluded Side}} \\
        \cline{2-7}
        & \multicolumn{1}{c}{CD\textcolor{blue}{$\downarrow$}} & \multicolumn{1}{c}{P2F\textcolor{blue}{$\downarrow$}} & \multicolumn{1}{c|}{Normal\textcolor{blue}{$\downarrow$}} & \multicolumn{1}{c}{CD\textcolor{blue}{$\downarrow$}} & \multicolumn{1}{c}{P2F\textcolor{blue}{$\downarrow$}} & \multicolumn{1}{c}{Normal\textcolor{blue}{$\downarrow$}} \\
        \hline
        \hline
        PIFu \cite{saito2019pifu} & 1.381   & 1.376  & 1.993  & 1.421  & 1.463  & 2.347 \\ 
        ICON \cite{xiu2022icon} & 1.555   & 1.575 & 2.010  & 1.589   & 1.727 & 2.364 \\ 
        ECON \cite{xiu2022econ}  & 1.634   & 1.702   & 1.884 & 1.618   & 1.755   & 2.283 \\ 
        Our HaP & \textbf{0.960} & \textbf{1.070}   & \textbf{1.450} & \textbf{1.043}  & \textbf{1.206} & \textbf{1.780} \\ 
        \hline
        \color{gray}{IntegratedPIFu} \cite{chan2022integratedpifu} & \color{gray}{1.310} & \color{gray}{1.438}  & \color{gray}{0.987}  & \color{gray}{1.264} & \color{gray}{1.252}  & \color{gray}{1.892}\\ 
        \color{gray}{Our HaP-I}       & \color{gray}{\textbf{1.150}} & \color{gray}{\textbf{1.246}} & \color{gray} {1.736} & \color{gray}{\textbf{1.218}} & \color{gray}{1.374}  & \color{gray}{1.948} \\
        \hline
        \color{gray}{2K2K} \cite{2k2k}  & \color{gray}{1.972} & \color{gray}{2.331}  & \color{gray}{3.688}  &\color{gray}{1.739} & \color{gray}{1.935}  & \color{gray}{3.919} \\ 
        \color{gray}{Our HaP-T} & \color{gray}{\textbf{0.907}} & \color{gray}{\textbf{1.001}} & \color{gray}{\textbf{1.464}} &  \color{gray}{\textbf{1.009}} & \color{gray}{\textbf{1.176}}  & \color{gray}{\textbf{1.766}}\\
        \bottomrule[1.0pt]
    \end{tabular}
    \label{teql2}
\end{table*}

\subsection{Comparisons with State-of-the-Art Methods} 

We primarily compared the proposed HaP with PIFu \cite{saito2019pifu}\footnote{Note that the PIFu method used in this comparison is implemented by Xiu \textit{et al}. \cite{xiu2022icon} for its better performance than the original implementation.}, ICON \cite{xiu2022icon} and ECON \cite{xiu2022econ}. Additionally, we compared HaP with IntegratedPIFu \cite{chan2022integratedpifu} and 2K2K \cite{2k2k}, which are proposed for processing images of resolutions 1k and 2k, respectively. \revise{(All methods are trained on Thuman2.)}
 Following Saito \textit{et al}.~\cite{saito2019pifu} and Xiu \textit{et al}.~\cite{xiu2022icon}, we evaluated the performance of different methods using three metrics: CD \revise{(Chamfer Distance)}, P2F \revise{(Point-to-Face)} distance, and normal difference. To compute CD, we sampled points uniformly on both the ground truth and reconstructed meshes and calculated the average bi-directional point-to-face distances. 
The P2F distance is calculated as the average distance from ground truth points to the reconstructed mesh. To compute the normal difference, we render the ground truth and reconstructed meshes from four fixed views using Pytorch3D, and calculate the L2 distances between them. This captures high-frequency geometric details. 

Generally, as listed in Table \ref{sotaresults}, our HaP demonstrates superior quantitative performance compared to PIFu \cite{saito2019pifu}, ICON \cite{xiu2022icon}, ECON \cite{xiu2022econ}, and 2K2K \cite{2k2k} on both Thuman3.0 and CityUHuman datasets by a large margin. And HaP can achieve competitive performance comparable to IntegratedPIFu \cite{chan2022integratedpifu}.
Specifically, compared with existing fully-implicit learning approaches \cite{saito2019pifu,xiu2022icon,chan2022integratedpifu}, as visualized in Fig. \ref{errormap}, HaP can correctly reconstruct poses and exhibits superior local geometric detail recovery. Additionally,  Fig. \ref{compare1} showcase some instances where PIFu \cite{saito2019pifu} and ICON \cite{xiu2022icon} encounter challenges, such as self-occlusion and loose clothing \revise{(marked with \textcolor{red}{RED} circle)}, while our HaP successfully addresses them.

\revise{Note that, IntegratedPIFu \cite{chan2022integratedpifu} typically demands high-resolution images of $1024 \times 1024$ as input,} resulting in extensive rendering times for preparing the training data of our HaP. To facilitate a quantitative comparison, we trained and tested HaP with the same scans but a lower spatial resolution, i.e., $512 \times 512$. This method is referred to as HaP-I. Despite the data difference, our HaP-I still outperforms IntegratedPIFu in terms of CD and P2F  on Thuman3.0, as reported in Table \ref{sotaresults}.

In addition to fully-implicit learning, recent studies have also explored fully explicit and hybrid processing pipelines, such as ECON \cite{xiu2022econ} and 2K2K \cite{2k2k}. Both of them aim to combine two \revise{part}s of a human body, whereas HaP aims to generate a human body based on a partial point cloud. While ECON effectively addresses the issue of loose clothing, it still faces challenges related to its rendering-based SMPL refinement method. This can introduce SMPL inaccuracies when ECON combines two partial 2.5D meshes, resulting in unexpected outcomes, as illustrated in Fig. \ref{compare1} \revise{(marked with \textcolor{red}{RED} circle)}. With our proposed SMPL rectification, HaP consistently outperforms ECON, both quantitatively and qualitatively.
Moreover, 2K2K \cite{2k2k} usually fails to reconstruct human bodies when the Openpose \cite{cao2017realtime} cannot detect all human keypoints due to self-occlusion. Therefore, omitting the failure cases of 2K2K, we conducted a re-evaluation of our HaP, denoted as HaP-T. 
The quantitative results listed in Table \ref{sotaresults}  indicate that our HaP-T \revise{outperforms 2K2K on} both the Thuman3.0 and CityUHuman datasets. It is worth mentioning that the visual comparative results shown in Fig. \ref{compare1} reveal the limb absence of the reconstructed surface at occlusion boundaries, \revise{We consider that 2D keypoints fail to provide accurate 3D human body structural information effectively.} \revise{(marked with \textcolor{red}{RED} circle)} Compared to 2K2K \cite{2k2k}, our HaP-T can accurately reconstruct the entire human body while maintaining reasonable surface quality.

\revise{Moreover, we further extended the quantitative evaluation by separately measuring the visible and occluded sides of the reconstructed meshes in Table \ref{teql2}, where HaP basically achieves large performance gains against all the competing methods, except for IntegratedPIFu on the Normal metric for the visible side as well as the P2F and Normal metrics on the occluded side. However, it is worth pointing out again that the input image for IntegratedPIFu is of higher resolution (i.e., $1024 \times 1024$), while ours is of $512 \times 512$.}

Finally, to assess the effectiveness of our HaP in \textit{real-world} scenarios, we gathered a collection of images featuring diverse clothing styles and poses from the internet to infer their full 3D structures. Fig. \ref{compareinthwildoniconecon} visually compares the results of our HaP with ICON and ECON, 
where it can be seen that HaP can correctly reconstruct the clothing details and poses, while ICON cannot reconstruct the loose clothing details and ECON fails to recover correct human poses. More results of our HaP are depicted in Fig. \ref{wild}, demonstrating commendable performance in realistic scenarios. \revisesec{We also refer readers to the \textbf{hapresultsvideodemo.mp4} in the supplementary material}.

\begin{figure*}[t]
    \centering
    \includegraphics[width=4.8in]{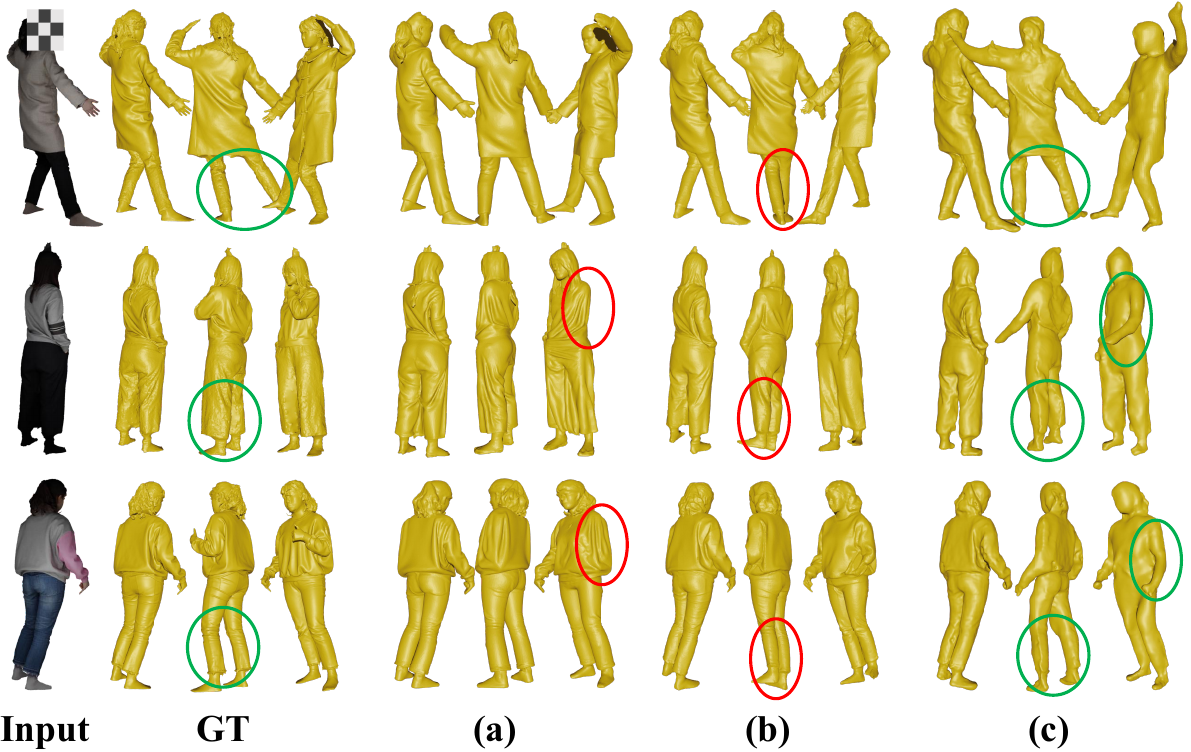}
     \caption{Comparison with generative models. (a) Hunyuan3D 2.0 \cite{zhao2025hunyuan3d}. (b) PSHuman \cite{li2024pshuman}. (c) Ours.}
    \label{minorvisual}
\end{figure*}

\begin{table}[h]
    \centering
    \caption{Quantitative comparisons of Hunyuan3D 2.0 \cite{zhao2025hunyuan3d}, PSHuman \cite{li2024pshuman} and our Hap on Thuman3.}
    \label{minortable}
    \begin{tabular}{c|ccc}
        \toprule
         & CD & P2F & Normal \\ \hline
        Hunyuan3D & 1.267 & 1.229 & 2.985  \\ \hline
        PSHuman & 1.045 &  1.171 & 2.386 \\ \hline
        HaP & \textbf{0.860} & \textbf{0.925} & \textbf{2.170} \\ \bottomrule
    \end{tabular}   
\end{table}
\subsection{Comparisons with 3D Generative Methods}

\revisesec{We also compared HaP with two 3D generative methods: Hunyuan3D 2.0 \cite{zhao2025hunyuan3d},  a general-purpose approach, and PSHuman \cite{li2024pshuman}, a human-centric method. 
Note that both models were trained with large-scale datasets, whereas HaP, specifically for generating 3D human bodies, was trained only with 500 scans from Thuman2.0 \cite{tao2021function4d}. As shown in Table~\ref{minortable}, HaP achieves the best overall quantitative performance.}

\revisesec{In addition to the numerical results, we conducted a visual comparison to further analyze the strengths and limitations of each method. As illustrated in Fig.~\ref{minorvisual}, PSHuman \cite{li2024pshuman} occasionally incorrectly generates closed-leg postures instead of the expected open-leg stance, highlighting its limitations in pose estimation. In contrast, HaP accurately preserves such structural details.}
\revisesec{Furthermore, Hunyuan3D 2.0 \cite{zhao2025hunyuan3d}
sometimes struggles to generate complete body parts, particularly under severe occlusions. 
This limitation suggests that relying solely on data-driven learning without incorporating strong human priors can hinder the generation of complete human bodies.}

\begin{table}[t]
    \centering
    \renewcommand\arraystretch{1.25}
    \setlength{\tabcolsep}{4.5pt}
    \caption{\revise{Comparison of different SMPL rectification methods. Thuman2 is used for evaluation.}}
    \begin{tabular}{lccc|cc|c}
        \toprule
        & Depth & Normal & PaMIR  & only P2F & w/o Keypoint & Ours \\
        \midrule
        CD \textcolor{blue}{$\downarrow$} & 1.664 & 1.502 & 1.243  & 1.056 & 0.823 & 0.818\\
        P2F \textcolor{blue}{$\downarrow$} & 1.714 & 1.548 & 1.319  & 1.100 & 0.869 & 0.860\\
        Normal \textcolor{blue}{$\downarrow$} & 2.354 & 2.271 & 2.064  & 1.929 & 1.559 & 1.554 \\
        \bottomrule
    \end{tabular}
    \label{r2c2c3}
\end{table}

\begin{figure*}[h]
    \centering
    \includegraphics[width=1.0\linewidth]{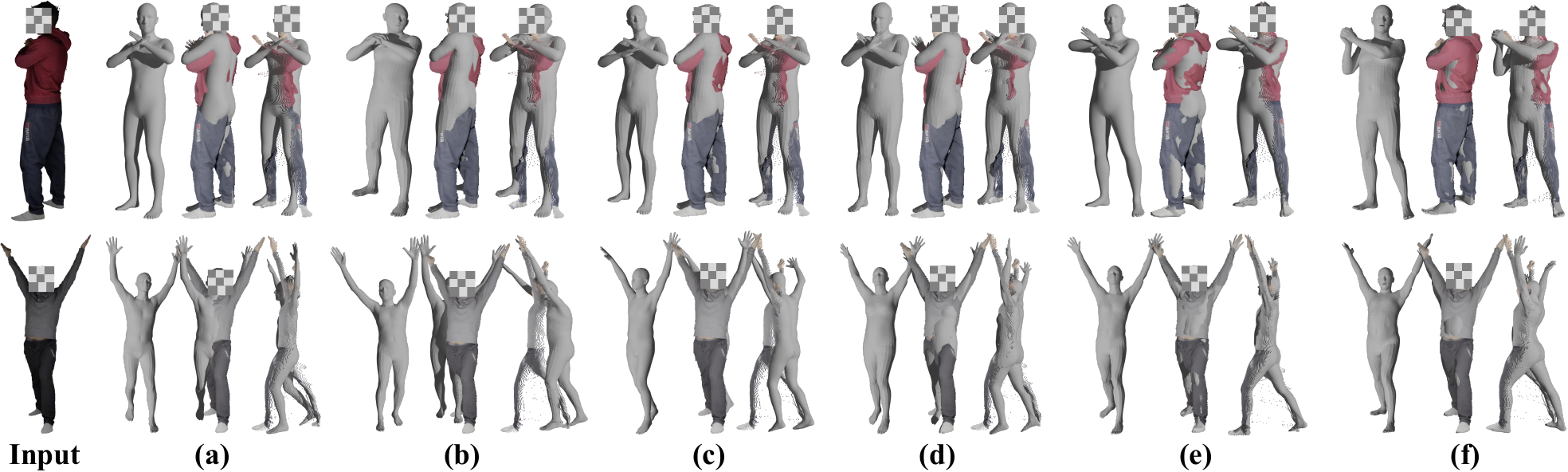}
    \caption{Visual comparisons of different SMPL rectification methods. (a) \revise{I}nitial predicted. (b) \revise{D}epth-based rendering. (c) \revise{N}ormal-based rendering. (d) \revise{P}aMIR. (e) \revise{O}urs. (f) \revise{G}round Truth.}
    \label{fr2f11}
\end{figure*}
\subsection{Analysis of SMPL Rectification}
\label{analysisonsmpl}
The accuracy of an SMPL model is pivotal for the precise reconstruction of the 3D human body. Nonetheless, as illustrated in Fig. \ref{fr2f11}, SMPL models directly predicted from RGB images frequently exhibit incorrect poses when viewed in 3D space. Xiu et al. \cite{xiu2022icon} sought to tackle this problem by introducing a rendering-based optimization method. However, this approach still possesses certain limitations.

\subsubsection{Defects of rendering-based optimization}
Xiu et al. \cite{xiu2022icon} employed a technique involving the rendering normal maps of SMPL models; subsequently, they attempted to optimize the SMPL models by minimizing the L1 distance between these SMPL normal maps and estimated normal maps of RGB images. However, this approach shows some inherent limitations. The pixels in normal maps are often not perfectly aligned, making the L1 distance an imperfect metric for quantifying the dissimilarities between the two normal maps. Consequently, the optimized SMPL models may exhibit inaccuracies and unnatural poses.
Despite the incorporation of human prior information, ICON's performance doesn't exhibit significant improvements when compared to PIFu, as illustrated in Table \ref{sotaresults}. This highlights the negative impact of imprecise SMPL models on the overall performance.

\revise{In addition to optimizing SMPL models with normal maps, we further experimented with depth maps. The qualitative and quantitative comparisons are presented in Fig. \ref{fr2f11} and Table \ref{r2c2c3}, respectively.}

\begin{figure}[t]
    \centering
    \includegraphics[width=3.1in]{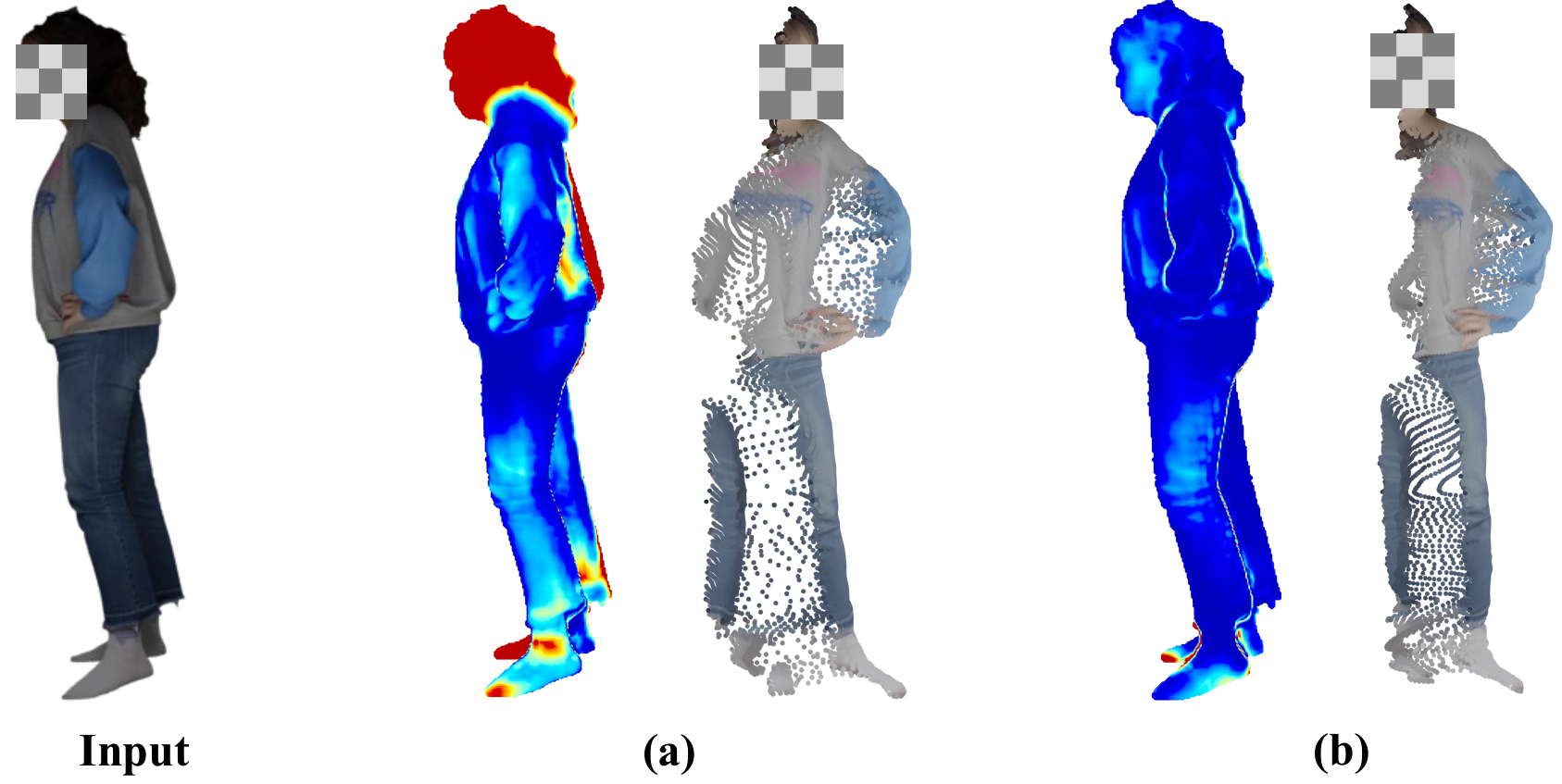}
     \caption{The error maps and partial point clouds of depths predicted by (a) P3Depth \cite{patil2022p3depth} and (b) MIM \cite{xie2023revealing}.}
    \label{comparedepth}
\end{figure}

\begin{figure}[t]
    \centering
    \includegraphics[width=3.1in]{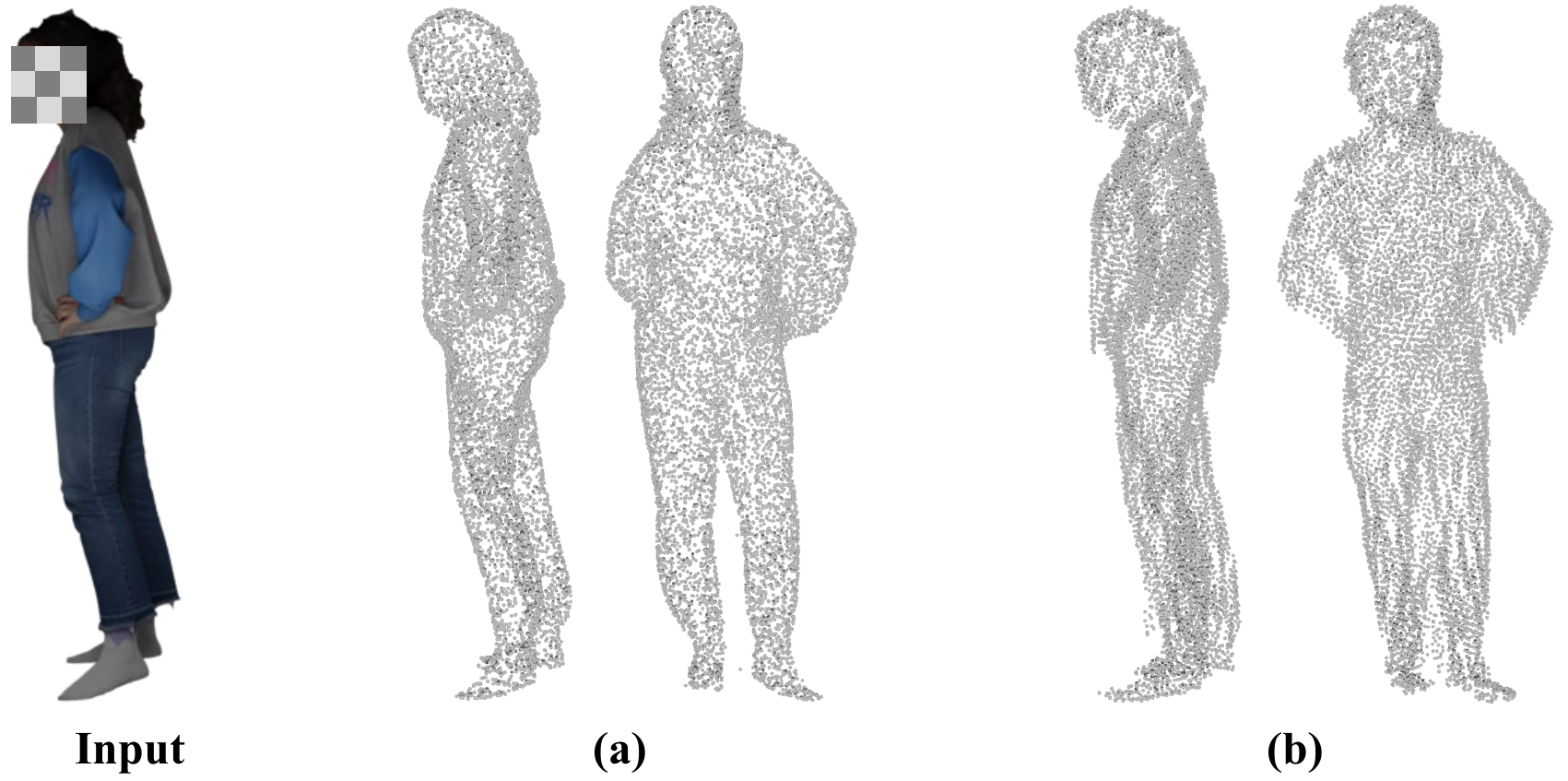}
     \caption{The point clouds generated by (a) MIM+PDR and (b) MIM+VRCNet.} 
    \label{pccompare}
\end{figure}
\begin{figure}[t]
    \centering
    \includegraphics[width=3.1in]{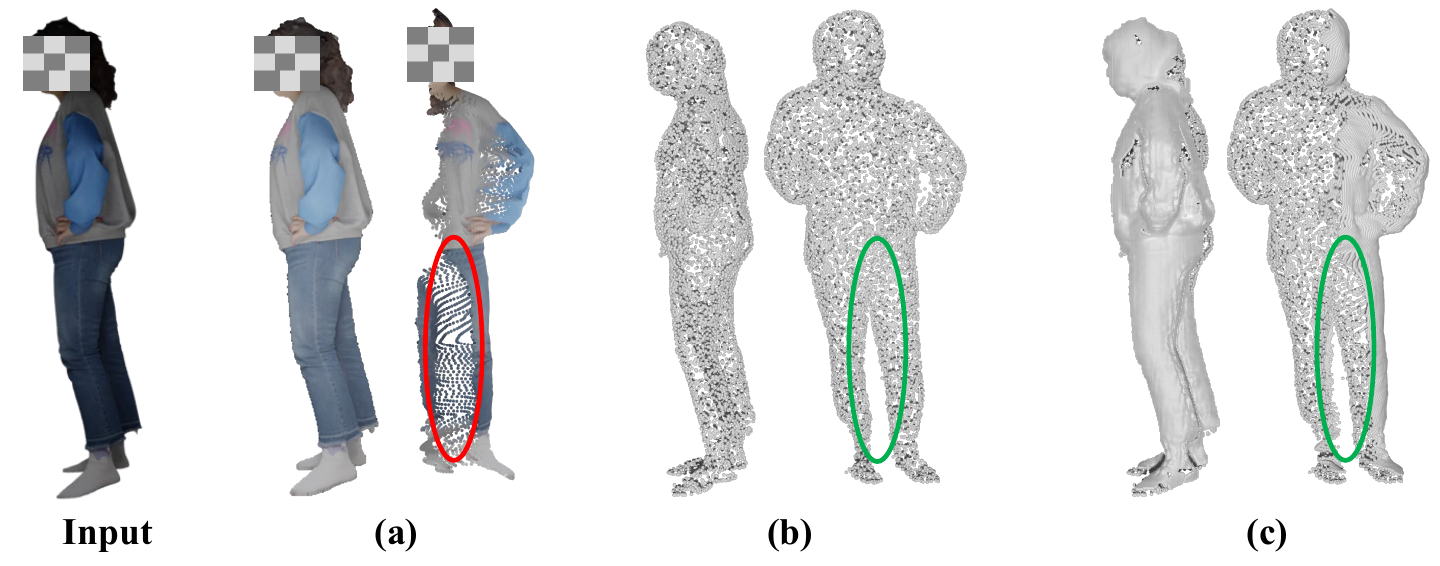}
     \caption{\revise{(a) Predicted partial point cloud. (b) Diffusion-based generated point cloud. (c) Point cloud after depth replacement.}} 
    \label{newfigureremvoenoisepoints}
\end{figure}

\subsubsection{The effectiveness of optimization in 3D space}
\revise{Several methods \cite{wang2021locally, bhatnagar2020combining} have \revise{evaluated} the efficacy of registering SMPL models within the implicit 3D space. In the context of monocular human reconstruction, PaMIR~\cite{zheng2021pamir} also optimizes the SMPL models in the implicit 3D space by minimizing the occupancy loss to make the SMPL align with the predicted implicit function, as shown in Table \ref{r2c2c3} and Fig. \ref{fr2f11}, it achieves better quantitative and qualitative results than rendering-based methods. }

\revise{HaP  minimizes the distances between the SMPL model and the partial point cloud, facilitated by $\texttt{P2F}(\mathcal{P},\mathcal{S})$.} 
\revise{Sometimes, the nearest point found by P2F for the partial point cloud belongs to the invisible part of the SMPL model. This can lead to the invisible surface of SMPL being drawn nearer to the partial points, making the SMPL model unnatural, as visually illustrated in Fig. 12 (a).}
Furthermore, Fig. \ref{biandehebubiande} (b) \revise{underlines} the significance of 2D keypoint regularization. In its absence, the keypoints of the human body may be displaced to incorrect positions. By implementing our meticulously designed approach, the refined SMPL model shows a more accurate and natural pose as presented in Fig. \ref{biandehebubiande} (c). Results in Table \ref{r2c2c3} also demonstrate the necessity of our regularizations in the SMPL rectification module.

Our proposed SMPL rectification method outperforms the rendering methods by accurately registering with the partial point cloud. This leads to better performance for HaP using our efficient and effective SMPL rectification method.

\begin{figure}[t]
    \centering
    \includegraphics[width=3.5in]{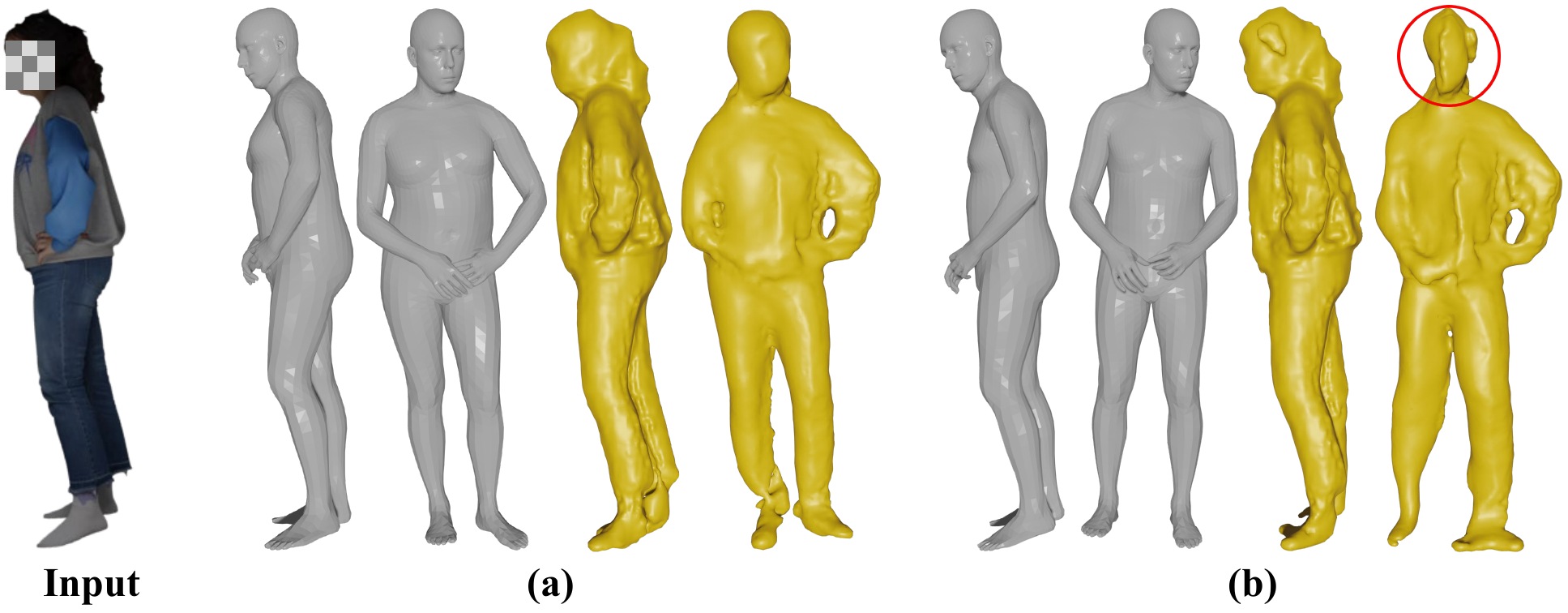}
     \caption{\revise{Visual comparison between the results by HaP (a) with and (b) without the SMPL rectification module. We present two views of the SMPL models and the reconstructed surfaces, respectively.}
     }
    \label{wosmplrefine}
\end{figure}

\begin{figure}[t]
    \centering
    \includegraphics[width=3.5in]{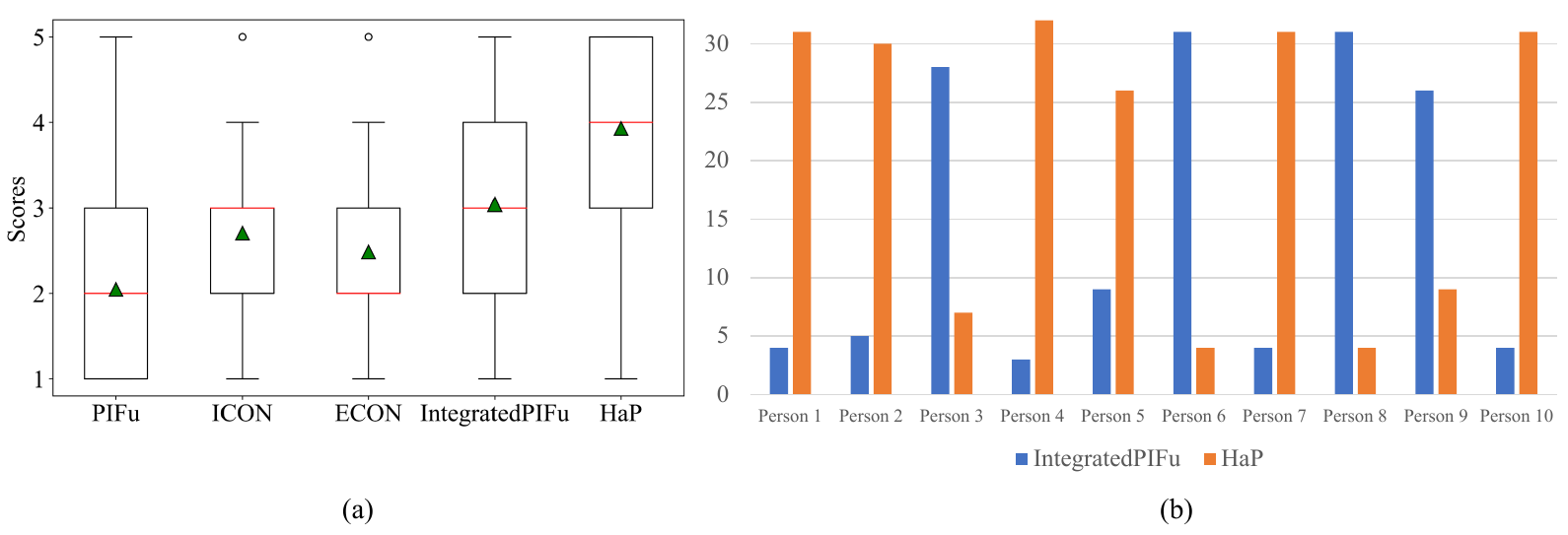}
     \caption{\revise{(a) The score boxplot of different methods (\textcolor{green}{Green Triangle} is the average scores, \textcolor{red}{Red Line} is the mid-value.) (b) Times of IntegratedPIFu and HaP being selected as the most realistic per person. } }
    \label{subjectivehuman}
\end{figure}

 \begin{table*}[t]
    \centering
    \renewcommand\arraystretch{1.25}
    \setlength{\tabcolsep}{10.0pt}
    \caption{The ablation studies on different modules of HaP. ``SR" and ``DR" are the SMPL rectification and depth replacement modules, respectively. \revise{Thuman3.0 is used for evaluation.} The row highlighted in \textcolor{blue}{blue} corresponds to our final HaP. \revise{``$\downarrow$" means the lower, the better.}}
    \begin{tabular}{c|c|c|c|c|c|c|c|c}
        \toprule[1.0pt]
        \multirow{2}{*}{Depth} & \multirow{2}{*}{Point} & \multirow{2}{*}{SMPL} & \multirow{2}{*}{SR} & \multicolumn{2}{c|}{Refinement Stage} & \multirow{2}{*}{CD \textcolor{blue}{$\downarrow$}} & \multirow{2}{*}{P2F \textcolor{blue}{$\downarrow$}} & \multirow{2}{*}{Normal \textcolor{blue}{$\downarrow$}} \\
         &  &  &  & DR   & $\texttt{PNet}_{\Theta_2}$   & &&  \\
        \hline
        \hline
        \texttt{P3Depth} & \texttt{PNet} & \texttt{PyMAF} & \Checkmark&\Checkmark  &\Checkmark & 0.946 & 1.015 & 2.301\
        \\
        \texttt{MIM} & \texttt{VRCNet} & \texttt{PyMAF} &  \Checkmark&\Checkmark   &\Checkmark& 1.643 & 1.275 & 9.352 \\
         \texttt{MIM} & \texttt{PNet} & \texttt{PIXIE} &  \Checkmark& \Checkmark  &\Checkmark& 0.883 & 0.944 & 2.203 \\
         \texttt{MIM} & \texttt{PNet} & \texttt{PyMAF} &  \XSolidBrush &\Checkmark   &\Checkmark& 1.051 & 1.079 & 2.622 \\
         \texttt{MIM} & \texttt{PNet} & \texttt{PyMAF} & \Checkmark&\XSolidBrush  & \Checkmark& 0.877 & 0.944 & 7.082 \\
         \texttt{MIM} & \texttt{PNet} & \texttt{PyMAF} & \Checkmark& \Checkmark  &\XSolidBrush & {0.958} & {0.978} & {7.437}\\
        \textcolor{blue}{\texttt{MIM}} & \textcolor{blue}{\texttt{PNet}} & \textcolor{blue}{\texttt{PyMAF}} & \textcolor{blue}{\Checkmark}& \textcolor{blue}{\Checkmark} &\textcolor{blue}{\Checkmark} & \textbf{0.860} & \textbf{0.925} & \textbf{2.170}\\
        \bottomrule[1.0pt]
    \end{tabular}
    \label{ablation}
\end{table*}

\begin{figure*}[t]
    \centering
    \setlength{\abovecaptionskip}{0.1cm}
    \includegraphics[width=6.8in]{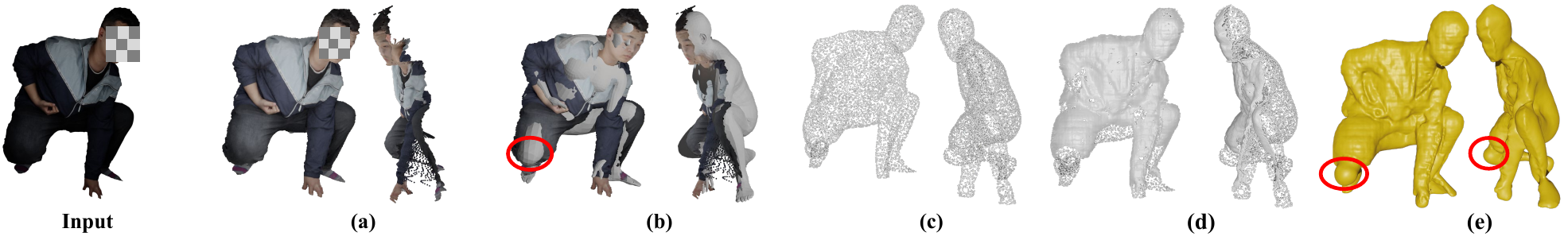}
    \caption{Illustration of the limitation resulted from the flawed SMPL prediction. (a) \revise{D}epth-\revise{inferred} point cloud. (b) \revise{R}efined SMPL model. (c) \revise{G}enerated point cloud. (d) \revise{R}eplaced point cloud. (e) \revise{R}econstructed surface. }
    \label{wrongsmpl}
\end{figure*}
 
\subsection{Ablation Studies}
We conducted comprehensive ablation studies on Thuman3.0 to gain a better understanding of the four modules encompassed in our HaP. For the depth estimation module, the conditioned diffusion-based generation module, and the SMPL estimation module, we considered three alternative methods:  P3Depth \cite{patil2022p3depth},  VRCNet \cite{pan2021variational}, and PIXIE \cite{feng2021collaborative}, respectively.

\subsubsection{Depth estimation} Note that MIM \cite{xie2023revealing} achieves more accurate depth on Thuman3.0 than P3Depth \cite{patil2022p3depth}, i.e., 0.0158 vs. 0.0189 in terms of RMSE. From the $1^{st}$ row of Table \ref{ablation}, as expected, a better depth estimation method can lead to better reconstruction performance.

However, as visualized in Fig. \ref{comparedepth}, 
\revise{both of them contain many noisy points}, which are further removed by our conditioned diffusion-based generation module.

\subsubsection{Conditioned diffusion-based generation} 
We replaced PNet \cite{lyu2021conditional} with VRCNet \cite{pan2021variational}, a point cloud completion network, to \revise{evaluate} the effectiveness of the diffusion-based generation network. 
As illustrated in Fig. \ref{pccompare},  VRCNet generates highly noisy point clouds, which adversely affects the accuracy of the depth replacement module and leads to poor performance of HaP, as indicated in the $2^{nd}$ row of Table \ref{ablation}, with a higher normal error. Moreover, our diffusion-based point cloud generation is able to effectively remove the noisy points produced by depth estimation (\revise{see the noisy points between legs in Fig. \ref{newfigureremvoenoisepoints}}).

\subsubsection{SMPL estimation and rectification} Compared to PyMAF \cite{zhang2021pymaf},  PIXIE \cite{feng2021collaborative} \revise{(the $3^{rd}$ row)} slightly under-performs in terms of CD and Normal metrics.
As listed in the 4$^{th}$ row of Table \ref{ablation}, although removing the SMPL rectification module does not drastically reduce the numerical results, \revise{the head looks rather weird and unnatural,} as exemplified in Fig. \ref{wosmplrefine}.

\revise{Moreover, we conducted experiments by replacing the SMPL models used in ICON and ECON with our HaP-rectified SMPL models. The modified methods are denoted as \textbf{ICON-S} and \textbf{ECON-S}.  As shown in Table~\ref{teql3}, a comparison between ICON with ICON-S and ECON with ECON-S reveals that HaP-refined SMPL models can improve the performance of ICON and ECON, demonstrating the effectiveness of SMPL rectification. Besides, Table~\ref{teql3} shows that HaP consistently outperforms ICON-S and ECON-S, underscoring the effectiveness of our explicit diffusion-based framework over the implicit function models utilized by ICON and ECON. }
\subsubsection{Refinement stage} As listed in the $5^{th}$ row of Table \ref{ablation},  although HaP without the depth replacement strategy achieves satisfactory performance in terms of CD and P2F, there is a significant increase in the Normal error, demonstrating \revise{its lack of effectiveness} in recovering surface details.

\revise{Furthermore, we assess the impact of omitting the displacement-based refinement module, $\texttt{PNet}{\Theta_2}$, which improves the point cloud's geometric details by learning point-wise displacements. As shown in the  $6^{th}$ and  $7^{th}$ rows of Table \ref{ablation}, the substantial performance drop across all metrics when $\texttt{PNet}{\Theta_2}$ is removed highlights its essential role in producing high-quality results.}

\begin{table}[h]
    \centering
    \renewcommand\arraystretch{1.25}
    \setlength{\tabcolsep}{10.0pt}
    \caption{Quantitative comparisons of ICON, ICON-S, ECON, ECON-S, and our Hap on Thuman3 when the same SMPL pose parameters obtained by our HaP were adopted by all methods.}
    \label{teql3}
    \begin{tabular}{c|ccc}
        \toprule
         & CD & P2F & Normal \\ \hline
        \textcolor{gray}{ICON} & \textcolor{gray}{1.306 }& \textcolor{gray}{1.395} &\textcolor{gray}{2.731 }\\ 
        ICON-S & 0.982 & 1.054 & 2.484 \\ \hline
        \textcolor{gray}{ECON} & \textcolor{gray}{1.467} &  \textcolor{gray}{1.524} & \textcolor{gray}{2.877} \\ 
        ECON-S & 0.970 &  1.034& 2.447 \\ \hline
        HaP & \textbf{0.860} & \textbf{0.925} & \textbf{2.170} \\ \bottomrule
    \end{tabular}   
\end{table}

\subsection{\revise{Perceptual Study}}
\label{sec:subjective}
We conducted \revise{two perceptual} evaluations to quantitatively compare various methods. It's important to note that 2K2K \cite{2k2k} was excluded from this evaluation, as it typically struggles to reconstruct the human body in cases where OpenPose \cite{cao2017realtime} fails to detect all keypoints. 

\revise{In the first perceptual evaluation, the questionnaire (\revisesec{Perceptual\_Evaluation\_1.pdf in the supplementary material}) consists of ten questions. Each question presented an RGB image and six videos. The first video in each question displayed the ground truth scanned model, while the remaining five videos corresponded to five different methods. The order of the methods was randomized across the ten questions. Participants were asked to rate the methods on a scale from 1 to 5 based on their similarity to the real model. For instance, if method A was judged as the most accurate, it would receive a score of 5; if it was considered the least accurate, it would receive a score of 1. An intuitive guide for scoring was provided (\revisesec{Intuitive\_Figure.pdf in the supplementary material}). Participants were not filtered via catch trials, and no special training was provided. The evaluation focused on aspects such as posture accuracy, limb completeness, clothing style reconstruction, rationality of the non-visible side, and surface smoothness. As shown in Fig. \ref{subjectivehuman} (a), HaP achieved the highest mean score. }

\revise{In the second perceptual evaluation (\revisesec{Perceptual\_Evaluation\_2.pdf in the supplementary material}), participants were shown the same videos and asked to select the best method between IntegratedPIFu and HaP, as they had obtained the highest scores in the first evaluation. Fig.~\ref{subjectivehuman} (b) shows that HaP was selected as the best method in six out of the ten samples.}

\subsection{Discussion on Limitations}
{\textbf{Depth Estimation.}  While we assume that depth estimation methods generally produce accurate depth maps for human bodies, we acknowledge that these methods can occasionally fail. Since our approach relies on depth estimation, we recognize that suboptimal results may occur when depth estimation is inaccurate.}

{\textbf{SMPL Rectification.}} HaP is \revise{also} constrained by the possibility of imperfect SMPL predictions. While our SMPL rectification method offers improved efficiency compared to rendering-based techniques, there are situations where it may not effectively rectify the SMPL models, even when the estimated depth maps are accurate. This can lead to suboptimal reconstruction results. For example, as illustrated in Fig. \ref{wrongsmpl}, HaP faces challenges in aligning the SMPL model's knee with the partial point cloud, resulting in a noticeable difference in the reconstruction outcome.

{\textbf{Smooth Occluded Side.} Another limitation is the overly smooth geometry on the occluded side. Predicting occluded geometry is particularly challenging due to the lack of occluded information. In this study, we did not utilize occluded side normal or depth map predictions, leaving the occluded areas unsupervised, whereas other methods~\cite{xiu2022econ,xiu2022icon,chan2022integratedpifu,2k2k} use. In future work, we plan to explore methods for incorporating appropriate supervision for the occluded side to achieve more realistic and natural geometry.}

\revise{\textbf{Loop-Closure Artifact.}  Some scans in the CityUHuman dataset may exhibit “loop-closure” artifacts. This occurs because, during the scanning process, the scanner had to be repositioned to complete the human body scan. This process, which typically lasts a few minutes, could result in small movements by the volunteer, causing repetitions in certain areas.}

\section{Conclusion} \label{sectionconclusion}

We presented HaP, a novel point-based learning pipeline for single-view 3D human reconstruction. After coarsely inferring 3D information from the 2D image domain, our approach is featured by explicit point cloud manipulation, generation, and refinement in the original 3D geometric space. Compared with previous mainstream implicit field-based approaches, the proposed HaP is highly flexible, generalizable, robust, and capable of modeling fine-level geometric details. Extensive experiments demonstrate the superiority of our HaP over the current state-of-the-art methods. \revise{It is foreseeable that with the continuous development of various foundational modules and architectures such as depth estimation and diffusion models, the overall performance of our HaP pipeline will also improve progressively.}

\normalem

{
\bibliographystyle{ieee_fullname}
\bibliography{egbib}

\begin{thebibliography}{10}\itemsep=-1pt

\bibitem{githubGitHubDanielgatisrembg}
Rembg.
\newblock \url{https://github.com/danielgatis/rembg}.

\bibitem{achlioptas2018learning}
Panos Achlioptas, Olga Diamanti, Ioannis Mitliagkas, and Leonidas Guibas.
\newblock Learning representations and generative models for 3d point clouds.
\newblock In {\em International Conferenceon Machine Learning}, pages 40--49,
  2018.

\bibitem{alldieck2022photorealistic}
Thiemo Alldieck, Mihai Zanfir, and Cristian Sminchisescu.
\newblock Photorealistic monocular 3d reconstruction of humans wearing
  clothing.
\newblock In {\em IEEE/CVF Computer Vision and Pattern Recognition Conference},
  pages 1506--1515, 2022.

\bibitem{bhatnagar2020combining}
Bharat~Lal Bhatnagar, Cristian Sminchisescu, Christian Theobalt, and Gerard
  Pons-Moll.
\newblock Combining implicit function learning and parametric models for 3d
  human reconstruction.
\newblock In {\em European Conference on Computer Vision}, 2020.

\bibitem{bogo2016keep}
Federica Bogo, Angjoo Kanazawa, Christoph Lassner, Peter Gehler, Javier Romero,
  and Michael~J Black.
\newblock {Keep it SMPL}: Automatic estimation of 3d human pose and shape from
  a single image.
\newblock In {\em European Conference on Computer Vision}, pages 561--578,
  2016.

\bibitem{cai2020learning}
Ruojin Cai, Guandao Yang, Hadar Averbuch-Elor, Zekun Hao, Serge Belongie, Noah
  Snavely, and Bharath Hariharan.
\newblock Learning gradient fields for shape generation.
\newblock In {\em European Conference on Computer Vision}, pages 364--381,
  2020.

\bibitem{cao2022bilateral}
Xu Cao, Hiroaki Santo, Boxin Shi, Fumio Okura, and Yasuyuki Matsushita.
\newblock Bilateral normal integration.
\newblock In {\em European Conference on Computer Vision}, pages 552--567,
  2022.

\bibitem{cao2017realtime}
Zhe Cao, Tomas Simon, Shih-En Wei, and Yaser Sheikh.
\newblock Realtime multi-person 2d pose estimation using part affinity fields.
\newblock In {\em IEEE/CVF Computer Vision and Pattern Recognition Conference},
  pages 7291--7299, 2017.

\bibitem{chan2022integratedpifu}
Kennard~Yanting Chan, Guosheng Lin, Haiyu Zhao, and Weisi Lin.
\newblock {IntegratedPIFu}: Integrated pixel aligned implicit function for
  single-view human reconstruction.
\newblock In {\em European Conference on Computer Vision}, pages 328--344,
  2022.

\bibitem{chen2023beyond}
Weihua Chen, Xianzhe Xu, Jian Jia, Hao Luo, Yaohua Wang, Fan Wang, Rong Jin,
  and Xiuyu Sun.
\newblock {Beyond appearance: a semantic controllable self-supervised learning
  framework for human-centric visual tasks}.
\newblock In {\em IEEE/CVF Computer Vision and Pattern Recognition Conference},
  pages 15050--15061, 2023.

\bibitem{chen2019learning}
Zhiqin Chen and Hao Zhang.
\newblock Learning implicit fields for generative shape modeling.
\newblock In {\em IEEE/CVF Computer Vision and Pattern Recognition Conference},
  pages 5939--5948, 2019.

\bibitem{corona2023structured}
Enric Corona, Mihai Zanfir, Thiemo Alldieck, Eduard~Gabriel Bazavan, Andrei
  Zanfir, and Cristian Sminchisescu.
\newblock Structured 3d features for reconstructing controllable avatars.
\newblock In {\em IEEE/CVF Computer Vision and Pattern Recognition Conference},
  pages 16954--16964, 2023.

\bibitem{eigen2014depth}
David Eigen, Christian Puhrsch, and Rob Fergus.
\newblock Depth map prediction from a single image using a multi-scale deep
  network.
\newblock {\em Advances in Neural Information Processing Systems}, 27, 2014.

\bibitem{fan2017point}
Haoqiang Fan, Hao Su, and Leonidas~J Guibas.
\newblock A point set generation network for 3d object reconstruction from a
  single image.
\newblock In {\em IEEE/CVF Computer Vision and Pattern Recognition Conference},
  pages 605--613, 2017.

\bibitem{alphapose}
Hao-Shu Fang, Jiefeng Li, Hongyang Tang, Chao Xu, Haoyi Zhu, Yuliang Xiu,
  Yong-Lu Li, and Cewu Lu.
\newblock {Alphapose: Whole-body regional multi-person pose estimation and
  tracking in real-time}.
\newblock {\em IEEE Transactions on Pattern Analysis and Machine Intelligence},
  45(6):7157--7173, 2022.

\bibitem{feng2021collaborative}
Yao Feng, Vasileios Choutas, Timo Bolkart, Dimitrios Tzionas, and Michael~J
  Black.
\newblock Collaborative regression of expressive bodies using moderation.
\newblock In {\em International Conference on 3D Vision}, pages 792--804, 2021.

\bibitem{gabeur2019moulding}
Valentin Gabeur, Jean-S{\'e}bastien Franco, Xavier Martin, Cordelia Schmid, and
  Gregory Rogez.
\newblock Moulding humans: Non-parametric 3d human shape estimation from single
  images.
\newblock In {\em IEEE/CVF international conference on computer vision}, pages
  2232--2241, 2019.

\bibitem{2k2k}
Sang-Hun Han, Min-Gyu Park, Ju~Hong Yoon, Ju-Mi Kang, Young-Jae Park, and
  Hae-Gon Jeon.
\newblock {High-fidelity 3d human digitization from single 2k resolution
  images}.
\newblock In {\em IEEE/CVF Computer Vision and Pattern Recognition Conference},
  pages 12869--12879, 2023.

\bibitem{he2021arch++}
Tong He, Yuanlu Xu, Shunsuke Saito, Stefano Soatto, and Tony Tung.
\newblock {ARCH++}: Animation-ready clothed human reconstruction revisited.
\newblock In {\em IEEE/CVF International Conference on Computer Vision}, pages
  11046--11056, 2021.

\bibitem{heo2018monocular}
Minhyeok Heo, Jaehan Lee, Kyung-Rae Kim, Han-Ul Kim, and Chang-Su Kim.
\newblock Monocular depth estimation using whole strip masking and
  reliability-based refinement.
\newblock In {\em European Conference on Computer Vision}, pages 36--51, 2018.

\bibitem{ho2020denoising}
Jonathan Ho, Ajay Jain, and Pieter Abbeel.
\newblock Denoising diffusion probabilistic models.
\newblock {\em Advances in Neural Information Processing Systems},
  33:6840--6851, 2020.

\bibitem{hornauer2022gradient}
Julia Hornauer and Vasileios Belagiannis.
\newblock Gradient-based uncertainty for monocular depth estimation.
\newblock In {\em European Conference on Computer Vision}, pages 613--630,
  2022.

\bibitem{huang2024humannorm}
Xin Huang, Ruizhi Shao, Qi Zhang, Hongwen Zhang, Ying Feng, Yebin Liu, and Qing
  Wang.
\newblock {HumanNorm}: Learning normal diffusion model for high-quality and
  realistic 3d human generation.
\newblock In {\em IEEE/CVF Computer Vision and Pattern Recognition Conference},
  pages 4568--4577, 2024.

\bibitem{huang2020arch}
Zeng Huang, Yuanlu Xu, Christoph Lassner, Hao Li, and Tony Tung.
\newblock {ARCH}: Animatable reconstruction of clothed humans.
\newblock In {\em IEEE/CVF Computer Vision and Pattern Recognition Conference},
  pages 3093--3102, 2020.

\bibitem{jafarian2021learning}
Yasamin Jafarian and Hyun~Soo Park.
\newblock Learning high fidelity depths of dressed humans by watching social
  media dance videos.
\newblock In {\em IEEE/CVF Computer Vision and Pattern Recognition Conference},
  pages 12753--12762, 2021.

\bibitem{kazhdan2013screened}
Michael Kazhdan and Hugues Hoppe.
\newblock Screened poisson surface reconstruction.
\newblock {\em ACM Transactions on Graphics}, 32:1--13, 2013.

\bibitem{kirillov2023segment}
Alexander Kirillov, Eric Mintun, Nikhila Ravi, Hanzi Mao, Chloe Rolland, Laura
  Gustafson, Tete Xiao, Spencer Whitehead, Alexander~C Berg, Wan-Yen Lo, et~al.
\newblock Segment anything.
\newblock In {\em International Conference on Computer Vision}, pages
  4015--4026, 2023.

\bibitem{kocabas2020vibe}
Muhammed Kocabas, Nikos Athanasiou, and Michael~J Black.
\newblock {VIBE: Video inference for human body pose and shape estimation}.
\newblock In {\em IEEE/CVF Computer Vision and Pattern Recognition Conference},
  pages 5253--5263, 2020.

\bibitem{lee2019big}
Jin~Han Lee, Myung-Kyu Han, Dong~Wook Ko, and Il~Hong Suh.
\newblock From big to small: Multi-scale local planar guidance for monocular
  depth estimation.
\newblock {\em arXiv preprint arXiv:1907.10326}, 2019.

\bibitem{li2024pshuman}
Peng Li, Wangguandong Zheng, Yuan Liu, Tao Yu, Yangguang Li, Xingqun Qi,
  Mengfei Li, Xiaowei Chi, Siyu Xia, Wei Xue, et~al.
\newblock Pshuman: Photorealistic single-view human reconstruction using
  cross-scale diffusion.
\newblock {\em arXiv preprint arXiv:2409.10141}, 2024.

\bibitem{liang2015deep}
Xiaodan Liang, Si Liu, Xiaohui Shen, Jianchao Yang, Luoqi Liu, Jian Dong, Liang
  Lin, and Shuicheng Yan.
\newblock {Deep human parsing with active template regression}.
\newblock {\em IEEE Transactions on Pattern Analysis and Machine Intelligence},
  37(12):2402--2414, 2015.

\bibitem{lin2021end}
Kevin Lin, Lijuan Wang, and Zicheng Liu.
\newblock {End-to-end human pose and mesh reconstruction with transformers}.
\newblock In {\em IEEE/CVF Computer Vision and Pattern Recognition Conference},
  pages 1954--1963, 2021.

\bibitem{lin2022learning}
Siyou Lin, Hongwen Zhang, Zerong Zheng, Ruizhi Shao, and Yebin Liu.
\newblock Learning implicit templates for point-based clothed human modeling.
\newblock In {\em European Conference on Computer Vision}, pages 210--228,
  2022.

\bibitem{liu2021swin}
Ze Liu, Yutong Lin, Yue Cao, Han Hu, Yixuan Wei, Zheng Zhang, Stephen Lin, and
  Baining Guo.
\newblock {Swin transformer: Hierarchical vision transformer using shifted
  windows}.
\newblock In {\em International Conference on 3D Vision}, pages 10012--10022,
  2021.

\bibitem{SMPL:2015}
Matthew Loper, Naureen Mahmood, Javier Romero, Gerard Pons-Moll, and Michael~J.
  Black.
\newblock {SMPL}: A skinned multi-person linear model.
\newblock {\em ACM Transactions on Graphics}, 34:248:1--248:16, 2015.

\bibitem{lyu2021conditional}
Zhaoyang Lyu, Zhifeng Kong, Xudong Xu, Liang Pan, and Dahua Lin.
\newblock A conditional point diffusion-refinement paradigm for 3d point cloud
  completion.
\newblock {\em International Conference on Learning Representation}, 2022.

\bibitem{ma2021scale}
Qianli Ma, Shunsuke Saito, Jinlong Yang, Siyu Tang, and Michael~J Black.
\newblock {SCALE}: Modeling clothed humans with a surface codec of articulated
  local elements.
\newblock In {\em IEEE/CVF Computer Vision and Pattern Recognition Conference},
  pages 16082--16093, 2021.

\bibitem{ma2021power}
Qianli Ma, Jinlong Yang, Siyu Tang, and Michael~J Black.
\newblock The power of points for modeling humans in clothing.
\newblock In {\em International Conference on 3D Vision}, pages 10974--10984,
  2021.

\bibitem{mescheder2019occupancy}
Lars Mescheder, Michael Oechsle, Michael Niemeyer, Sebastian Nowozin, and
  Andreas Geiger.
\newblock {Occupancy Networks}: Learning 3d reconstruction in function space.
\newblock In {\em IEEE/CVF Computer Vision and Pattern Recognition Conference},
  pages 4460--4470, 2019.

\bibitem{ming2021deep}
Yue Ming, Xuyang Meng, Chunxiao Fan, and Hui Yu.
\newblock Deep learning for monocular depth estimation: A review.
\newblock {\em Neurocomputing}, 438:14--33, 2021.

\bibitem{natsume2019siclope}
Ryota Natsume, Shunsuke Saito, Zeng Huang, Weikai Chen, Chongyang Ma, Hao Li,
  and Shigeo Morishima.
\newblock {SiCloPe}: Silhouette-based clothed people.
\newblock In {\em IEEE/CVF Computer Vision and Pattern Recognition Conference},
  pages 4480--4490, 2019.

\bibitem{nichol2022point}
Alex Nichol, Heewoo Jun, Prafulla Dhariwal, Pamela Mishkin, and Mark Chen.
\newblock Point-e: A system for generating 3d point clouds from complex
  prompts.
\newblock {\em arXiv preprint arXiv:2212.08751}, 2022.

\bibitem{pan2021variational}
Liang Pan, Xinyi Chen, Zhongang Cai, Junzhe Zhang, Haiyu Zhao, Shuai Yi, and
  Ziwei Liu.
\newblock {Variational relational point completion network}.
\newblock In {\em IEEE/CVF Computer Vision and Pattern Recognition Conference},
  pages 8524--8533, 2021.

\bibitem{park2019deepsdf}
Jeong~Joon Park, Peter Florence, Julian Straub, Richard Newcombe, and Steven
  Lovegrove.
\newblock {DeepSDF}: Learning continuous signed distance functions for shape
  representation.
\newblock In {\em IEEE/CVF Computer Vision and Pattern Recognition Conference},
  pages 165--174, 2019.

\bibitem{patil2022p3depth}
Vaishakh Patil, Christos Sakaridis, Alexander Liniger, and Luc Van~Gool.
\newblock {P3Depth: Monocular depth estimation with a piecewise planarity
  prior}.
\newblock In {\em IEEE/CVF Computer Vision and Pattern Recognition Conference},
  pages 1610--1621, 2022.

\bibitem{patni2024ecodepth}
Suraj Patni, Aradhye Agarwal, and Chetan Arora.
\newblock {ECoDepth}: Effective conditioning of diffusion models for monocular
  depth estimation.
\newblock In {\em IEEE/CVF Computer Vision and Pattern Recognition Conference},
  pages 28285--28295, 2024.

\bibitem{SMPL-X:2019}
Georgios Pavlakos, Vasileios Choutas, Nima Ghorbani, Timo Bolkart, Ahmed A.~A.
  Osman, Dimitrios Tzionas, and Michael~J. Black.
\newblock Expressive body capture: 3d hands, face, and body from a single
  image.
\newblock In {\em IEEE/CVF Computer Vision and Pattern Recognition Conference},
  pages 10975--10985, 2019.

\bibitem{peebles2023scalable}
William Peebles and Saining Xie.
\newblock Scalable diffusion models with transformers.
\newblock In {\em IEEE/CVF International Conference on Computer Vision}, pages
  4195--4205, 2023.

\bibitem{poole2023dreamfusion}
Ben Poole, Ajay Jain, Jonathan~T. Barron, and Ben Mildenhall.
\newblock Dreamfusion: Text-to-3d using 2d diffusion.
\newblock In {\em International Conference on Learning Representation}, 2023.

\bibitem{qi2017pointnet++}
Charles~Ruizhongtai Qi, Li Yi, Hao Su, and Leonidas~J Guibas.
\newblock {PointNet++: Deep hierarchical feature learning on point sets in a
  metric space}.
\newblock {\em Advances in Neural Information Processing Systems}, 30, 2017.

\bibitem{qian2023magic123}
Guocheng Qian, Jinjie Mai, Abdullah Hamdi, Jian Ren, Aliaksandr Siarohin, Bing
  Li, Hsin-Ying Lee, Ivan Skorokhodov, Peter Wonka, Sergey Tulyakov, et~al.
\newblock Magic123: One image to high-quality 3d object generation using both
  2d and 3d diffusion priors.
\newblock {\em International Conference on Learning Representation}, 2024.

\bibitem{qian2020pugeo}
Yue Qian, Junhui Hou, Sam Kwong, and Ying He.
\newblock {PUGeo-Net: A geometry-centric network for 3D point cloud
  upsampling}.
\newblock In {\em European conference on computer vision}, pages 752--769.
  Springer, 2020.

\bibitem{ren2023geoudf}
Siyu Ren, Junhui Hou, Xiaodong Chen, Ying He, and Wenping Wang.
\newblock {GeoUDF: Surface reconstruction from 3d point clouds via
  geometry-guided distance representation}.
\newblock In {\em International Conference on Computer Vision}, pages
  14214--14224, 2023.

\bibitem{ren2024measuring}
Siyu Ren, Junhui Hou, Xiaodong Chen, Hongkai Xiong, and Wenping Wang.
\newblock Measuring the discrepancy between 3d geometric models using
  directional distance fields.
\newblock {\em arXiv preprint arXiv:2401.09736}, 2024.

\bibitem{ren2024xcube}
Xuanchi Ren, Jiahui Huang, Xiaohui Zeng, Ken Museth, Sanja Fidler, and Francis
  Williams.
\newblock Xcube: Large-scale 3d generative modeling using sparse voxel
  hierarchies.
\newblock In {\em IEEE/CVF Computer Vision and Pattern Recognition Conference},
  pages 4209--4219, 2024.

\bibitem{saito2019pifu}
Shunsuke Saito, Zeng Huang, Ryota Natsume, Shigeo Morishima, Angjoo Kanazawa,
  and Hao Li.
\newblock {PIFu}: Pixel-aligned implicit function for high-resolution clothed
  human digitization.
\newblock In {\em International Conference on 3D Vision}, pages 2304--2314,
  2019.

\bibitem{saito2020pifuhd}
Shunsuke Saito, Tomas Simon, Jason Saragih, and Hanbyul Joo.
\newblock {PIFuHD}: Multi-level pixel-aligned implicit function for
  high-resolution 3d human digitization.
\newblock In {\em IEEE/CVF Computer Vision and Pattern Recognition Conference},
  pages 84--93, 2020.

\bibitem{saxena2008make3d}
Ashutosh Saxena, Min Sun, and Andrew~Y Ng.
\newblock {Make3d: Learning 3d scene structure from a single still image}.
\newblock {\em IEEE Transactions on Pattern Analysis and Machine Intelligence},
  31(5):824--840, 2008.

\bibitem{smith2019facsimile}
David Smith, Matthew Loper, Xiaochen Hu, Paris Mavroidis, and Javier Romero.
\newblock Facsimile: Fast and accurate scans from an image in less than a
  second.
\newblock In {\em IEEE/CVF international conference on computer vision}, pages
  5330--5339, 2019.

\bibitem{su2022deepcloth}
Zhaoqi Su, Tao Yu, Yangang Wang, and Yebin Liu.
\newblock {DeepCloth}: Neural garment representation for shape and style
  editing.
\newblock {\em IEEE Transactions on Pattern Analysis and Machine Intelligence},
  45(2):1581--1593, 2022.

\bibitem{sun2023dreamcraft3d}
Jingxiang Sun, Bo Zhang, Ruizhi Shao, Lizhen Wang, Wen Liu, Zhenda Xie, and
  Yebin Liu.
\newblock Dreamcraft3d: Hierarchical 3d generation with bootstrapped diffusion
  prior.
\newblock {\em International Conference on Learning Representation}, 2024.

\bibitem{tan2020self}
Feitong Tan, Hao Zhu, Zhaopeng Cui, Siyu Zhu, Marc Pollefeys, and Ping Tan.
\newblock Self-supervised human depth estimation from monocular videos.
\newblock In {\em IEEE/CVF Computer Vision and Pattern Recognition Conference},
  pages 650--659, 2020.

\bibitem{tang2019neural}
Sicong Tang, Feitong Tan, Kelvin Cheng, Zhaoyang Li, Siyu Zhu, and Ping Tan.
\newblock A neural network for detailed human depth estimation from a single
  image.
\newblock In {\em International Conference on 3D Vision}, pages 7750--7759,
  2019.

\bibitem{tang2023high}
Sicong Tang, Guangyuan Wang, Qing Ran, Lingzhi Li, Li Shen, and Ping Tan.
\newblock High-resolution volumetric reconstruction for clothed humans.
\newblock {\em ACM Transactions on Graphics}, 2023.

\bibitem{tang2022warpinggan}
Yingzhi Tang, Yue Qian, Qijian Zhang, Yiming Zeng, Junhui Hou, and Xuefei Zhe.
\newblock {WarpingGAN: Warping multiple uniform priors for adversarial 3D point
  cloud generation}.
\newblock In {\em IEEE/CVF Computer Vision and Pattern Recognition Conference},
  pages 6397--6405, 2022.

\bibitem{wang2020normalgan}
Lizhen Wang, Xiaochen Zhao, Tao Yu, Songtao Wang, and Yebin Liu.
\newblock {NormalGAN}: Learning detailed 3d human from a single rgb-d image.
\newblock In {\em European Conference on Computer Vision}, pages 430--446,
  2020.

\bibitem{wang2021locally}
Shaofei Wang, Andreas Geiger, and Siyu Tang.
\newblock Locally aware piecewise transformation fields for 3d human mesh
  registration.
\newblock In {\em IEEE/CVF Computer Vision and Pattern Recognition Conference},
  pages 7639--7648, 2021.

\bibitem{xia2020generating}
Zhihao Xia, Patrick Sullivan, and Ayan Chakrabarti.
\newblock Generating and exploiting probabilistic monocular depth estimates.
\newblock In {\em IEEE/CVF Computer Vision and Pattern Recognition Conference},
  pages 65--74, 2020.

\bibitem{xie2023revealing}
Zhenda Xie, Zigang Geng, Jingcheng Hu, Zheng Zhang, Han Hu, and Yue Cao.
\newblock Revealing the dark secrets of masked image modeling.
\newblock In {\em IEEE/CVF Computer Vision and Pattern Recognition Conference},
  pages 14475--14485, 2023.

\bibitem{xiong2022pifu}
Zhangyang Xiong, Dong Du, Yushuang Wu, Jingqi Dong, Di Kang, Linchao Bao, and
  Xiaoguang Han.
\newblock {PIFu for the Real World}: A self-supervised framework to reconstruct
  dressed human from single-view images.
\newblock {\em arXiv preprint arXiv:2208.10769}, 2022.

\bibitem{xiu2022econ}
Yuliang Xiu, Jinlong Yang, Xu Cao, Dimitrios Tzionas, and Michael~J Black.
\newblock {ECON: Explicit Clothed humans Obtained from Normals}.
\newblock {\em IEEE/CVF Computer Vision and Pattern Recognition Conference},
  2023.

\bibitem{xiu2022icon}
Yuliang Xiu, Jinlong Yang, Dimitrios Tzionas, and Michael~J Black.
\newblock {ICON}: implicit clothed humans obtained from normals.
\newblock In {\em IEEE/CVF Computer Vision and Pattern Recognition Conference},
  pages 13286--13296, 2022.

\bibitem{tao2021function4d}
Tao Yu, Zerong Zheng, Kaiwen Guo, Pengpeng Liu, Qionghai Dai, and Yebin Liu.
\newblock {Function4D}: Real-time human volumetric capture from very sparse
  consumer rgbd sensors.
\newblock In {\em IEEE/CVF Computer Vision and Pattern Recognition Conference},
  2021.

\bibitem{zhang2021pymaf}
Hongwen Zhang, Yating Tian, Xinchi Zhou, Wanli Ouyang, Yebin Liu, Limin Wang,
  and Zhenan Sun.
\newblock {PyMAF}: 3d human pose and shape regression with pyramidal mesh
  alignment feedback loop.
\newblock In {\em International Conference on 3D Vision}, pages 11446--11456,
  2021.

\bibitem{zhao2025hunyuan3d}
Zibo Zhao, Zeqiang Lai, Qingxiang Lin, Yunfei Zhao, Haolin Liu, Shuhui Yang,
  Yifei Feng, Mingxin Yang, Sheng Zhang, Xianghui Yang, et~al.
\newblock Hunyuan3d 2.0: Scaling diffusion models for high resolution textured
  3d assets generation.
\newblock {\em arXiv preprint arXiv:2501.12202}, 2025.

\bibitem{zheng2021pamir}
Zerong Zheng, Tao Yu, Yebin Liu, and Qionghai Dai.
\newblock {PaMIR}: Parametric model-conditioned implicit representation for
  image-based human reconstruction.
\newblock {\em IEEE transactions on pattern analysis and machine intelligence},
  44(6):3170--3184, 2021.

\bibitem{zhou20213d}
Linqi Zhou, Yilun Du, and Jiajun Wu.
\newblock 3d shape generation and completion through point-voxel diffusion.
\newblock In {\em International Conference on 3D Vision}, pages 5826--5835,
  2021.

\end{thebibliography}
}

\begin{IEEEbiography}[{\includegraphics[width=1in,height=1.25in,clip,keepaspectratio]{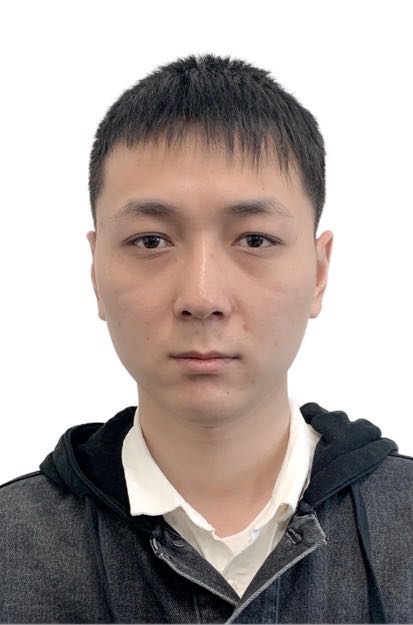}}]{Yingzhi Tang} received his B.S. degree in Computer Science and M.S. degree in Information Engineering from Xidian University, Xi'an, China, in 2018 and 2021 respectively. Currently, he is a Ph.D. student (2021-present) under the Department of Computer Science, City University of Hong Kong, HKSAR. His research focuses on deep learning-based point cloud processing, 3D human reconstruction for various 3D vision tasks.
\end{IEEEbiography}

\begin{IEEEbiography}[{\includegraphics[width=1in,height=1.25in,clip,keepaspectratio]{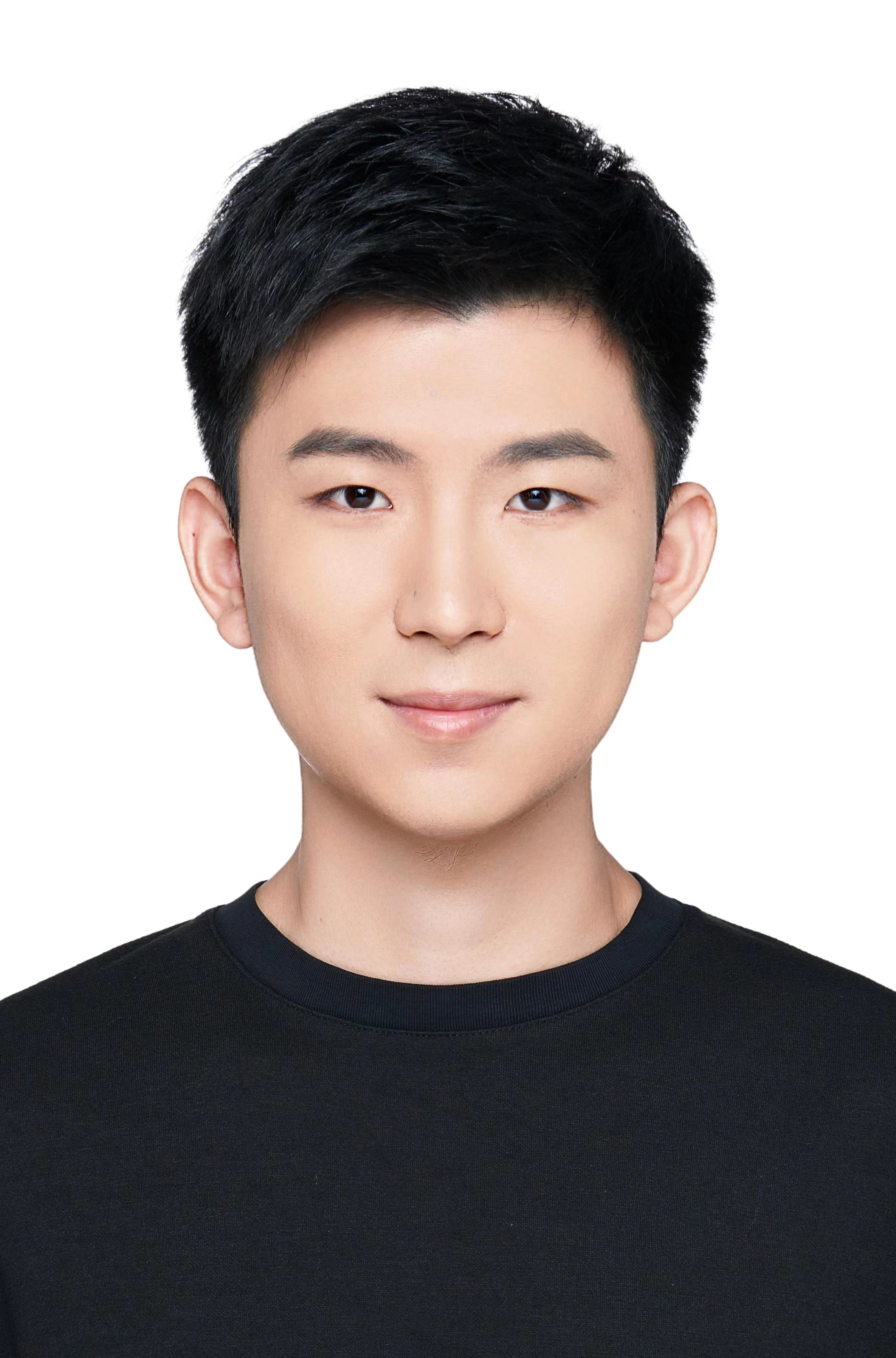}}]{Qijian Zhang} received the B.S. degree in Electronic Information Science and Technology from Beijing Normal University, Beijing, China, in 2019, and then the Ph.D. degree in Computer Science from City University of Hong Kong, Hong Kong SAR, in 2024. He is currently an applied researcher in the TiMi L1 Studio of Tencent Games, focusing on 3D generative models and AI for games.
\end{IEEEbiography}

\begin{IEEEbiography}[{\includegraphics[width=1in,height=1.25in,clip,keepaspectratio]{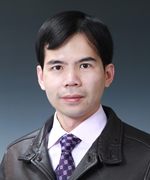}}]
{Yebin Liu} (Member, IEEE) received the BE degree from the Beijing University of Posts and Telecommunications, China, in 2002, and the PhD degree from the Department of Automation, Tsinghua University, Beijing, China, in 2009. He is currently a professor with Tsinghua University. He was a research fellow with the Computer Graphics Group of the Max Planck Institute for Informatik, Germany, in 2010. His research areas include computer vision, computer graphics, and computational photography.
\end{IEEEbiography}

\begin{IEEEbiography}[{\includegraphics[width=1in,height=1.25in,clip,keepaspectratio]{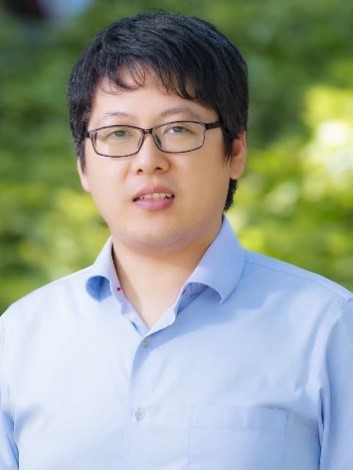}}]{Junhui Hou}
is an Associate Professor with the Department of Computer Science, City University of Hong Kong. He holds a B.Eng. degree in information engineering (Talented Students Program) from the South China University of Technology, Guangzhou, China (2009), an M.Eng. degree in signal and information processing from Northwestern Polytechnical University, Xi’an, China (2012), and a Ph.D. degree from the School of Electrical and Electronic Engineering, Nanyang Technological University, Singapore (2016). His research interests are multi-dimensional visual computing.

Dr. Hou received the Early Career Award from the Hong Kong Research Grants Council in 2018 and the NSFC Excellent Young Scientists Fund in 2024. He has served or is serving as an Associate Editor for \textit{IEEE Transactions on Visualization and Computer Graphics}, \textit{IEEE Transactions on Image Processing}, \textit{IEEE Transactions on Multimedia}, and \textit{IEEE Transactions on Circuits and Systems for Video Technology}.
\end{IEEEbiography}
 
\end{document}